%% file: main.tex
\documentclass[conference]{IEEEtran}
\IEEEoverridecommandlockouts
\usepackage{cite}
\usepackage{amsmath,amssymb,amsfonts}
\usepackage{algorithmic}
\usepackage{graphicx}
\usepackage{textcomp}
\usepackage{xcolor}
\usepackage{bm}
\usepackage{makecell}

\usepackage{booktabs} % For formal tables
\usepackage{enumitem}
\usepackage{bbding}
\usepackage{graphicx}
\usepackage{graphicx}
\usepackage{balance}
\usepackage{algorithmic}
\usepackage{subfigure}
\usepackage{multirow}
\usepackage{color}
\usepackage{colortbl}
\usepackage{latexsym}
\usepackage{epsfig}
\usepackage{bbm}
\usepackage[misc]{ifsym}
\usepackage{url}
\usepackage{graphicx}
\usepackage{boxedminipage}
\usepackage{pgfplots}
\usepackage{pgf-pie}

\newcommand{\grd}{\nabla}

\newcommand{\X}{\mathbf{X}}
\newcommand{\x}{\mathbf{x}}
\newcommand{\W}{\mathbf{W}}
\newcommand{\M}{\mathbf{M}}

\renewcommand{\S}{\mathbf{S}}

\newcommand{\pa}{\text{pa}}

\newcommand{\uppdelta}{\overline{\delta}}

\newcommand{\algo}{\textsc{LEAST} }

\definecolor{mygray}{gray}{.9}

\newtheorem{lemma}{Lemma}

\def\BibTeX{{\rm B\kern-.05em{\sc i\kern-.025em b}\kern-.08em
    T\kern-.1667em\lower.7ex\hbox{E}\kern-.125emX}}
\begin{document}

\title{Efficient and Scalable Structure Learning for Bayesian Networks: Algorithms and Applications\thanks{
\rule{0.95\linewidth}{0.5pt} Rong Zhu and Andreas Pfadler contributed equally to this work, and Ziniu Wu is the corresponding author.}
}

\author{\IEEEauthorblockN{Rong Zhu\IEEEauthorrefmark{1}, Andreas Pfadler\IEEEauthorrefmark{1}, Ziniu Wu\IEEEauthorrefmark{1}\IEEEauthorrefmark{2}, Yuxing Han\IEEEauthorrefmark{1}, Xiaoke Yang\IEEEauthorrefmark{1}, \\ Feng Ye\IEEEauthorrefmark{1}, Zhenping Qian\IEEEauthorrefmark{1}, Jingren Zhou\IEEEauthorrefmark{1} and Bin Cui\IEEEauthorrefmark{3}}
\IEEEauthorblockA{\IEEEauthorrefmark{1}\textit{Alibaba Group, China}, \IEEEauthorrefmark{2}\textit{Oxford University, UK}, \IEEEauthorrefmark{3}\textit{Peking University, China} \\
\IEEEauthorrefmark{1}\{\textsf{red.zr, andreaswernerober,  ziniu.wzn, yuxing.hyx, xiaoke.yxk, keven.yef,} \\ \textsf{zhengping.qzp, jingren.zhou}\}\textsf{@alibaba-inc.com}, 
\IEEEauthorrefmark{3}\textsf{bin.cui@pku.edu.cn}
}
}

\maketitle

\begin{abstract}
Structure Learning for Bayesian network (BN) is an important problem
with extensive research. It plays central roles in a wide variety of applications in Alibaba Group. However, existing structure learning algorithms suffer from considerable limitations in real-world applications due to their low efficiency and poor scalability. To resolve this, we propose a new structure learning algorithm \textsc{LEAST}, which comprehensively fulfills our business requirements as it attains high accuracy, efficiency and  scalability at the same time. The core idea of \textsc{LEAST} is to formulate the structure learning into a continuous constrained optimization problem, with a novel
differentiable constraint function measuring the acyclicity of the resulting graph. Unlike with existing work, our constraint function is built on the spectral radius of the graph and could be evaluated in near linear time w.r.t.~the graph node size. Based on it, \textsc{LEAST} can be efficiently implemented with low storage overhead. According to our benchmark evaluation, \textsc{LEAST} runs $1$--$2$ orders of magnitude faster than state-of-the-art method with comparable accuracy, and it is able to scale on BNs with up to hundreds of thousands of variables.
In our production environment, \textsc{LEAST} is deployed and serves for more than 20 applications with thousands of executions per day. 
We describe a concrete scenario in a ticket booking service in Alibaba, where \algo is applied to build a near real-time automatic anomaly detection and root error cause analysis system. We also show that \textsc{LEAST} unlocks the possibility of applying BN structure learning in new areas, such as large-scale gene expression data analysis and explainable recommendation system.
\end{abstract}

%\begin{IEEEkeywords}
%structure learning, Bayesian networks, acyclicity constraint, efficiency, scalability, business applications
%\end{IEEEkeywords}

\maketitle

\input{intro.tex}

\input{background.tex}

\input{method.tex}

\input{algorithm.tex}

\input{exp.tex}

\input{application.tex}

\section{Conclusions and Future Work}
Driven by a wide variety of business scenarios in Alibaba, we design \textsc{LEAST}, a highly accurate, efficient and scalable on structure learning algorithm for BN. \textsc{LEAST} is built upon a novel way to characterize graph acyclicity using an upper bound of the graph spectral radius. Benchmark evaluation exhibits that \textsc{LEAST} attains state-of-the-art performance in terms of accuracy and efficiency. It is able to scale on BNs with hundreds of thousands nodes. Moreover, \textsc{LEAST} has been deployed in our production systems to serve a wide range of applications.

This work makes the first attempt to apply theoretically appealing method (BN) to address complex problems at the intersection of machine learning and causality, which are previously considered intractable. We believe that it would attract more efforts on  applying structure learning for BN to problems involving large amount of random variables. We also hope to implement \textsc{LEAST} in a distributed environment to explore extremely large-scale applications.

\bibliography{bibliography.bib}
\bibliographystyle{plain}

\end{document}

%% file: intro.tex
% !TeX spellcheck = en_US

\section{Introduction}

\label{sec: intro}

Discovering, modeling and understanding \emph{causal} mechanisms behind natural phenomena are fundamental tasks in numerous scientific disciplines, 
ranging from physics to economics, from biology to sociology. Bayesian network (BN) is prominent example of probabilistic graphical models and has been recognized as a powerful and versatile tool for modeling causality. Each BN takes the form of a directed acyclic graph (DAG), where each node corresponds to an observed or hidden variable and each edge defines the causal dependency between two variables. By further specifying the conditional probability distributions based on the causal structure, one eventually obtains a joint probability distribution for the model variables. 

In real-world situations, running randomized controlled trials uncovering causal relationships for BN can be costly and time consuming. Sometimes, it is even impossible  due to ethical concerns. Thus, it has been of great interest to develop statistical methods to infer the structure of BN purely based on observed data. This problem, called \emph{structure learning}, has become a research hot spot in machine learning (ML) community. 
Traditional applications include gene expression data analysis~\cite{lee2019scaling, scanagatta2019survey}, error identification~\cite{wellen2012learning}, model factorization~\cite{goudet2018learning} and system optimization~\cite{tzoumas2011lightweight}, to name but a few. 

%Over the past two decades, probabilistic graphical models (PGMs) such as Bayesian Networks (BN) have been extensively deployed in production systems for monitoring, anomaly detection, root cause analysis and related application scenarios.

\smallskip
\noindent{\underline{\textbf{Application Scenarios.}}}
In Alibaba Group, a very large-scale company, structure learning for BN has been applied in a wide range of business scenarios, including but not limited to the areas of e-commerce, cloud computing and finance. In these areas, monitoring production systems is a representative application of BN structure learning.
Traditional monitoring methods require a significant amount of human resources to analyze and understand log information compiled from a wide variety of internal and external interfaces. Thus, monitoring results heavily rely on personal experience, which may be inaccurate and incomplete. Meanwhile, due to complex data pipelines, business logic and technical infrastructure, staff feedback may not be obtained in time, greatly affecting the usability of some applications such as ticket booking and cloud security. For example, in an airline ticket booking application, operation staff often needs from several hours to days to find  possible reasons causing booking errors.

To overcome these limitations, BN is a robust and interpretable tool
for automatic anomaly detection and fast root error cause analysis in monitoring production systems. Due to the complexity and scale of our systems, BN needs to be generated in a data-driven fashion without relying too much on expert input. Therefore, it is very necessary to design a scalable and efficient structure learning method for BN to fulfill our business requirements. 

\smallskip
\noindent{\underline{\textbf{Challenges.}}}
Learning the structure of a BN purely from data is a highly nontrivial
endeavour, and it becomes much more difficult for our concrete business applications. Existing structure learning algorithms for BNs are based on discrete combinatorial optimization or numerical continuous optimization. Combinatorial optimization algorithms~\cite{park2017bayesian, schmidt2007learning, chickering2002optimal, hesar2012structure, fu2013learning, gu2019penalized, scanagatta2016learning, heckerman1995learning, kuipers2014addendum, chickering1997efficient, bouckaert1993probabilistic} can only produce accurate results for BNs with tens of nodes. Their accuracy significantly declines in the larger regimes. Recently,~\cite{lachapelle2019gradient, lee2019scaling, yu2019dag, zheng2018dags, zheng2019learning} proposed a new paradigm by measuring the acyclicity of a graph in a numerical manner. In this paradigm, the structure learning problem is transformed into a constrained continuous optimization problem and can be solved using off-the-shelf gradient based methods. Therefore, this class of algorithms can be easily implemented in modern ML platforms and tend to produce accurate results. Yet, they suffer from efficiency and scalability issues. The time and space cost of computing the acyclicity metric are $O(d^3)$ and  $O(d^2)$ for a $d$-node graph~\cite{zheng2018dags, zhu2019causal}, respectively, which only scales well to BNs with up to a thousand of nodes. As our application scenarios often contain tens of thousands to millions of variables, no existing method is applicable. %Practically, we require a highly scalable and efficient structure learning algorithm for BNs.

\smallskip
\noindent{\underline{\textbf{Our Contributions.}}}
In this paper, we tackle the above challenges by proposing a new structure learning algorithm called \textsc{LEAST}. It comprehensively fulfills our business requirements as it simultaneously attains high accuracy, high time efficiency and superior scalability. We leverage the advantages of continuous optimization algorithms, while designing a new way around their known drawbacks. 
Specifically, we find a novel and equivalent function to measure the acyclicity of the resulting graph based on its \emph{spectral properties}. This new acyclicity constraint function is differentiable and much easier to compute. Based on it, \textsc{LEAST} can be efficiently implemented with low space overhead. Therefore, \textsc{LEAST} is able to learn structure of very large BN.
We have deployed \textsc{LEAST} in the production environment of Alibaba. It now serves more than 20 business applications and is executed thousands of times per day.

Our main contributions can be summarized as follows:
\begin{itemize}[leftmargin = *]
\item We propose \textsc{LEAST}, a scalable, efficient and accurate structure learning algorithm for BN. It is built upon a new 
acyclicity constraint, whose time and space computation cost is near linear w.r.t.~the graph node size. (Sections 3 and 4)

\item We conduct extensive experiments on benchmark datasets to verify the effectiveness, efficiency and scalability of \textsc{LEAST}. It attains a comparable result accuracy w.r.t.~the state-of-the-art method while speeding up the computation by $1$--$2$ orders of magnitude. Furthermore, it can scale to BNs with up to hundreds of thousands nodes. (Section~5)

\item We demonstrate the usage of \textsc{LEAST} in different applications, including a production task in ticket booking system, a large-scale gene expression data analysis application and a case study in building explainable recommendation systems. We believe that this work represents a first step towards unlocking structure learning for a wide range of  scenarios. (Section~6)
\end{itemize}

%% file: background.tex
% !TeX spellcheck = en_US

\section{Background}
\label{sec: bkg}

We first review some relevant concepts and background knowledge on BN and structure learning in this section.

\begin{figure*}[!t]
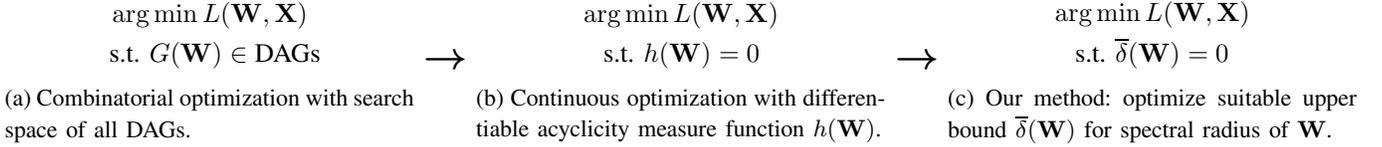

	\centering
	\begin{minipage}{0.3\linewidth}
		\begin{equation*}
		\begin{split}
		~ \arg & \min L(\W, \X) \\
		\text{s.t. } & G(\W) \in \text{DAGs}
		\end{split}
		\end{equation*}
		{\small (a) Combinatorial optimization with search space of all DAGs.}
	\end{minipage}
	{\Large \bm{$\rightarrow$}}
	\begin{minipage}{0.3\linewidth}
		\begin{equation*}
		\begin{split}
		\arg & \min L(\W, \X) \\
		\text{s.} & \text{t. } h(\W) = 0
		\end{split}
		\end{equation*}
		{\small (b) Continuous optimization with differentiable acyclicity measure function $h(\W)$.}
	\end{minipage}
	{\Large \bm{$\rightarrow$}}
	\begin{minipage}{0.3\linewidth}
		\begin{equation*}
		\begin{split}
		\arg & \min L(\W, \X) \\
		\text{s.} & \text{t. }\uppdelta(\W) = 0
		\end{split}
		\end{equation*}
		{\small (c) Our method: optimize suitable upper bound $\uppdelta(\W)$ for spectral radius of $\W$.}
	\end{minipage}
	\caption{\normalsize Different optimization paradigms of structure learning algorithms for BNs.}
	\label{Fig: OPA}
\end{figure*}

\smallskip
\noindent{\underline{\textbf{Bayesian Networks and Structural Equation Models.}}}
We suppose that in our problem of interest we can observe a $d$-dimensional vector\footnote{Unless otherwise statement, we assume all vectors to be column vectors.} $X \in \mathbb{R}^{d}$ of random variables. The random vector $X$ is assumed to be distributed according to some BN with DAG $G = (V, E)$, where $V$ and $E$ represent the set of nodes and edges, respectively. Thus, each node $i \in V$ exactly  corresponds to a random variable $X_{i}$ of $X$, and each edge $(i, j) \in E$ indicates some kind of causal relation from variable $X_{i}$ to $X_{j}$. Let $X_{\pa(i)}$ denote the set of parent random variables of $X_i$. Random variables in a BN model satisfy the \emph{first order Markov} property. That is, each variable $X_i$ is dependent on $X_{\pa(i)}$, but independent from all other random variables conditioned on $X_{\pa(i)}$. Therefore, one can decompose the high-dimensional joint probability density function (PDF) $p(X)$ of $X$ into a product of compact local mass functions as $p(X) = \prod\nolimits_{i = 1}^{d} p(X_i | X_{\pa(i)})$, where $p(X_i | X_{\pa(i)}) = p(X_i)$ if $X_i$ has no parents.

BNs admit an interesting interpretation: one may regard each variable $X_i$ as a stochastic function of $X_{\pa(i)}$, so that the causal relations encoded in a BN model naturally lead to a \emph{structural equation model} (SEM)~\cite{hoyer2009nonlinear, peters2017elements}. For $i = 1, 2, \dots, d$, let $n_i$ be a random noise variable and $f_i$ be a measurable function. In an SEM, we set $X_{i} =  f_{i}(X_{\pa(i)}) + n_i$,
to indicate the causal relation from parents $X_{\pa(i)}$ to the random variable $X_i$. The functions $f_{i} (\cdot)$ are modeled as some linear or non-linear transformations of $X_{\pa(i)}$ to which i.i.d. noise terms $n_i$ are added. Once $f_{i} (\cdot)$ and $n_i$ are suitably parameterized, an SEM may not only serve as a proxy to BN, but one may also \textit{learn} a BN via estimation of SEM parameters.

In this paper, we consider the widely used linear SEM (LSEM)~\cite{shimizu2006linear}. This means we set $X_{i} = w_{i}^{T} X$ where $w_i[j] = 0$ if $X_j$ is not a parent of $X_i$, and  noise terms $n_i$ are not restricted to be Gaussian.

\smallskip
\noindent{\underline{\textbf{Structure Learning Problem.}}}
Given a sample matrix $\X = [\x_{1}, \x_{2}, \dots, \x_{d}] \in \mathbb{R}^{n \times d}$ containing $n$ i.i.d.~ observations of the random vector $X$ assumed to follow a SEM model, the structure learning asks to output a BN $G$ which encodes the causal relations between all $X_i$. For the LSEM case, let the matrix $\W = [w_{1}, w_{2}, \dots, w_{d}]$. Learning the structure of $G$ is equivalent to finding the matrix $\W$ where node $i$ connects to node $j$ in $G$ iff $\W[i, j] \neq 0$. For clarity, we denote the graph induced by $\W$ as $G(\W)$. For a matrix $\W$, a decomposable loss function $L ( \cdot)$, such as the least squares or the negative log-likelihood on all nodes, may be used to evaluate how well the model corresponding to $\W$ fits the observed data. Hence, the structure learning problem is essentially solving the following program, which has been proven to be NP-Hard in ~\cite{chickering1996learning}:
\begin{equation}
\label{eq: slp-lsem}
\begin{split}
\arg\min\nolimits_{\W} L(\W, \X) & =  \frac{1}{n} \sum_{i = 1}^{d} L(\x_{i}, w_{i}^{T} \X), \\
\text{ subject to } & G(\W) \in \text{ DAGs}.
\end{split}
\end{equation}

\smallskip
\noindent{\underline{\textbf{Existing Structure Learning Algorithms.}}}
From Eq.~\eqref{eq: slp-lsem}, the key challenge in BN structure learning is how to enforce the acyclicity constraint for $G(\W)$. Historically this has been addressed using combinatorial optimization algorithms which directly explore the search space of all DAGs to optimize multiple kinds of loss scores~\cite{heckerman1995learning, kuipers2014addendum, chickering1997efficient, bouckaert1993probabilistic}. However, as the search space grows super-exponentially in the number of BN nodes, these methods can only scale to graphs with around ten nodes~\cite{robinson1977counting}. Later on, a large number of approximate combinatorial algorithms~\cite{park2017bayesian, schmidt2007learning, chickering2002optimal, heckerman1995learning, hesar2012structure, fu2013learning, gu2019penalized, scanagatta2016learning} have been proposed to speed up by some heuristic pruning rules. They improve the efficiency while suffer a significant decline in accuracy~\cite{ramsey2017million, scanagatta2015learning} in the large-scale regime. Besides, their scalability is still far from enough for many real-world applications~\cite{drton2017structure, lee2019scaling}.

As a breakthrough,~\cite{zheng2019learning} proposed a novel method called \textsc{NOTEARS} to recast structure learning into a continuous optimization problem by measuring the acyclicity in a numerical manner. As shown in Fig.~\ref{Fig: OPA}(b), $h(\W)$ is a smooth non-negative function evaluating the closeness of graph $G(\W)$ with DAG. $h(\W) = 0$ iff $G(\W)$ is exactly a DAG. Based on this, the structure learning problem is transformed into a constrained continuous optimization problem and can be solved using off-the-shelf gradient based optimization methods. Algorithms in this class~\cite{lachapelle2019gradient, lee2019scaling, yu2019dag, zheng2018dags} can generally handle a wide range of differentiable loss scores and tend to produce satisfying results. Yet, they suffer from efficiency and scalability issues since computing $h(\W)$ requires $O(d^3)$ time and $O(d^2)$ space for a $d$-node graph~\cite{zheng2018dags, zhu2019causal}.
According to our tests, \textsc{NOTEARS} can process at most a thousand of nodes on a NVIDA V100 GPU with 64GB RAM, requiring several hours until convergence.

\smallskip
\noindent{\underline{\textbf{Summary}}}
Existing structure learning algorithms for BNs suffer from considerable limitations w.r.t efficiency and scalability. None of them is applicable in our company's business scenarios such as root error cause analysis and recommendation systems containing tens of thousands to millions of variables. To this end, we explore a new path to design efficient and scalable structure learning methods fulfilling our practical requirements. In the following Section~3, we propose a new acyclicity constraint, which establishes a new paradigm of structure learning for BNs. The detailed algorithm and its applications are then presented in Section~4 and~6, respectively.

%% file: method.tex
% !TeX spellcheck = en_US

\section{New Acyclicity Constraint}
\label{sec: nco}

In this section, we address the key challenge of structure learning for BNs by proposing a novel acyclicity constraint. We present our fundamental idea in Section~\ref{sec: nco-paf}.
Details on the formulation are given in Section~\ref{sec: nco-nac}. Finally, in Section~\ref{sec: nco-cm} we present the actual computation scheme for the constraint.

\subsection{Preparations and Foundations}
\label{sec: nco-paf}
We first revisit the original constraint as presented in \cite{zheng2018dags}. Based on this, we then explain the key points of our work.

\smallskip
\noindent \underline{\textbf{Revisiting Existing Acyclicity Constraints.}}
Let $\S = \W \circ \W$ where $\circ$ is the Hadamard product. Then, $G$ is a DAG iff   
\begin{equation}
\label{eq: wdag}
h(\S) = \text{Tr}(e^{\S}) - d = 0, 
\end{equation}
where $e^{\S}$ is the matrix exponential of $\S$. 

We can regard $\S$ as a non-negative adjacency matrix of the graph $G$, where each positive element $\S[i, j]$ indicates an edge $(i, j)$ in $G$. For each $k \geq 1$ and every node $i$, $\S^{k}[i, i]$ is the sum of weights of all $k$-length cycles passing through node $i$ in $G$. Therefore, there exist some cycles in $G$ iff $\text{Tr}(\S^{k}) > 0$ for some $k \geq 1$. Since $e^{\S} = \sum_{i = 0}^{\infty} \frac{\S^{k}}{k!}$ where $\S^{0} = \mathbf{I}$, Eq.~\eqref{eq: wdag} certainly indicates that there exist no cycles in $G$. Later,~\cite{yu2019dag} relaxes this constraint to 
\vspace{-1em}
\begin{equation}
\label{eq: wdagnew}
g(\S) = \text{Tr}((\mathbf{I} + \S)^{d}) - d = \text{Tr} (\sum_{k = 1}^{d} {d \choose k} \S^{k}) = 0.
\end{equation}
For an acyclic graph $G$ Eq.~\eqref{eq: wdagnew} holds since a simple cycle in $G$ contains at most $d$ nodes. 

The acyclicity metrics $h(\S)$ or $g(\S)$ have two inherent drawbacks: 
1) costly operations such as matrix exponential or matrix multiplication, whose complexity is $O(d^3)$; 
and 2) costly storage of the dense matrix $e^{\S}$ or $(\mathbf{I} + \S)^{d}$ even though $\S$  is often sparse.
These two drawbacks fundamentally limit the efficiency and scalability of existing continuous optimization algorithms. To overcome the drawbacks, we need to design a new acyclicity metric for $\S$ with \emph{lightweight} computation and space overhead. 

\smallskip
\noindent \underline{\textbf{Fundamental Idea.}}
We try to characterize the acyclicity of graph using the \emph{spectral properties} of the matrix $S$.
Without loss of generality, let $\delta_{1}, \delta_{2}, \dots, \delta_{d}$ be the $d$ eigenvalues of matrix $\S$. Since $\S$ is non-negative, we have: 
1) $\delta_{i} \geq 0$ for all $1 \leq i \leq d$;
2) $\text{Tr}(\S) = \sum_{i = 1}^{d} \delta_{i}$;
and 3) $\delta_{1}^{k}, \delta_{2}^{k}, \dots, \delta_{d}^{k}$ are the $d$ eigenvalues of matrix $\S^{k}$ for all $k \geq 2$. The absolute value of the largest eigenvalue $\delta$ is called the \emph{spectral radius} of $\S$. By its definition, we obviously have $\sum_{k = 1}^{\infty} \text{Tr} (\S^{k}) = 0$ iff $\delta = 0$.
Therefore, the spectral radius can be used as a measure for acyclicity. 

Prior work~\cite{lee2019scaling} has used $\delta$ for the acyclicity constraint. However, computing an exact or approximate spectral radius also requires $O(d^{3})$ or $O(d^2)$ time, respectively.
To this end, instead of using $\delta$ itself, we try to utilize a suitable proxy by deriving an upper bound $\uppdelta$ of $\delta$, and then optimize Eq.~\eqref{eq: slp-lsem} by asymptotically decreasing $\uppdelta$ to a very small value. When $\uppdelta$ is small enough, $\delta$ will be close to $0$. As shown in Fig.~\ref{Fig: OPA}(c), this establishes a new paradigm of structure learning for BNs.

\smallskip
\noindent \underline{\textbf{Requirements on $\uppdelta$.}}
Obtaining an upper bound on the spectral radius is a longstanding mathematical problem and closely related to the well-known Perron-Frobenius theorem~\cite{chang2008perron}. However, finding a suitable $\uppdelta$ is a non-trivial task, which should satisfy the following requirements:

\begin{enumerate} [leftmargin = 2em]
\item[\textbf{R1:}] We can ensure acyclicity by using $\uppdelta$ as a proxy for $\delta$ during the optimization process, i.e., $\uppdelta$  is \textbf{\textit{consistent}} to the exact $\delta$.

\item[\textbf{R2:}] \textbf{\textit{Differentiability}} w.r.t.~$\S$ so that $\uppdelta$ can be optimized with off-the-shelf gradient based methods.
	
\item[\textbf{R3:}] $\uppdelta$ and its  gradient $\grd_{\S} \uppdelta$ should be \textbf{\textit{time-efficient}} to compute  without involving costly matrix operations.

\item[\textbf{R4:}] $\uppdelta$ and its  gradient $\grd_{\S} \uppdelta$ should be \textbf{\textit{space-efficient}} to compute without costly intermediate storage overhead.
\end{enumerate}

\subsection{Acyclicity Constraint Formulation}
\label{sec: nco-nac}

We formalize our proposed upper bound $\uppdelta$ in this subsection and show that it satisfies the above requirements. Given a matrix $\mathbf{A}$, let $r(\mathbf{A})$ and $c(\mathbf{A})$ be the vector of row sums and column sums of $\mathbf{A}$, respectively. Given a vector $v$ and any real value $\alpha$, let $v^{\alpha}$ be the resulting element-wise power vector. Our bound is derived in an iterative manner. Let $\S^{(0)} = \S = \W \circ \W$. 
For any $k \geq 0$ and $0 \leq \alpha \leq 1$, let $b^{(k)} = {(r(\S^{(k)}))}^{\alpha} \circ {{(c(\S^{(k)})})}^{1 - \alpha}$, $\mathbf{D}^{(k)}  = \text{Diag}(b^{(k)})$ and 
\begin{equation}
\label{eq: uppdelta}
\S^{(k + 1)} = ~ {(\mathbf{D}^{(k)})}^{-1} \S^{(k)} \mathbf{D}^{(k)}, 
\end{equation}
where we set ${(\mathbf{D}^{(k)})}^{-1} [i, i] = 0$ if $ \mathbf{D}^{(k)}[i, i] = 0$. We set the upper bound $\uppdelta^{(k)} = \sum_{i = 1}^{d} b^{(k)}[i]$. The following lemma states the correctness of this upper bound, following~\cite{Surhone2011Spectral}. Due to space limits, we omit all proofs in this version.

\begin{lemma}
\label{lem: uppdelta}
For any non-negative matrix $\S$, $k \geq 0$ and $0 \leq \alpha \leq 1$, the spectral radius $\delta$ of matrix $\S$ is no larger than $\uppdelta^{(k)}$.
\end{lemma}

In Eq.~\eqref{eq: uppdelta}, we apply a diagonal matrix transformation on $\S^{(k)}$ and $\uppdelta^{(k)}$ gradually approaches the exact $\delta$~\cite{Surhone2011Spectral}. 
In our experiments, we find that setting $k$ to a small number around $5$ is enough to ensure the accuracy of the results. The factor $\alpha$ is a hyper-parameter balancing the effects of $r(\S^{(k)})$ and $c(\S^{(k)})$. 
We set it closer to $0$ when values in $r(\S^{(k)})$ are much larger than those in $c(\S^{(k)})$ and vice versa so that the upper bound $\uppdelta^{(k)}$ will be smaller.

 \smallskip
\noindent \underline{\textbf{Consistency between $\uppdelta$ and $h(\S)$ and $g(\S)$.}}
Next, we verify one by one that our upper bound satisfies all stated requirements. At first, the following lemma establishes the consistency between the upper bound $\delta^{(k)}$ and the original acyclicity metrics $h(\S)$ and $g(\S)$. 

\begin{lemma}
\label{lem: consisybound}
For any non-negative matrix $\S$, $k \geq 0$ and any value $ \epsilon, \alpha \in (0, 1)$, 
if the upper bound $\uppdelta^{(k)} \leq \ln(\frac{\epsilon}{d} + 1)$, then
$h(\S) \leq \epsilon$ holds; if the upper bound $\uppdelta^{(k)} \leq \frac{1}{\alpha} \log_{d} \frac{\epsilon}{d^{2}}$, then $g(\S) \leq \epsilon$ holds.
\end{lemma}

By Lemma~\ref{lem: consisybound}, it follows that it is safe to use $\delta^{(k)}$ as a proxy for $h(\S)$ and $g(\S)$ since they will also decrease to a very small value when we optimize $\delta^{(k)}$. According to our benchmark evaluation results in Section~\ref{sec: exp}, the correlation between $\delta^{(k)}$ and $h(\S)$ often exceeds $0.9$. This indicates that R1 is satisfied.
For the other requirements (R2--R4), we reserve the examination of them for the following subsection.

\subsection{Efficient Computation of $\uppdelta$ and $\grd_{\S} \uppdelta$}
\label{sec: nco-cm}

In this section, we introduce a time and space efficient way to compute the upper bound $\uppdelta^{(k)}$ and its gradient $\grd_{\S} \uppdelta^{(k)}$. 

 \smallskip
\noindent \underline{\textbf{Computing $\uppdelta^{(k)}$.}}
Given a matrix $\mathbf{A}$ and a column vector $v$, let $\mathbf{A} \circ v$ (or $v \circ \mathbf{A}$) and $\mathbf{A} \circ v^{T}$ (or $v^{T} \circ \mathbf{A}$) denote the resulting matrix by multiplying $v$ onto each column of $\mathbf{A}$ and $v^{T}$ onto each row of $\mathbf{A}$, respectively.
The diagonal transformation in Eq.~\eqref{eq: uppdelta} can then be equivalently written as  
\begin{equation}
\label{eq: mulskd}
\S^{(k + 1)} = {(\mathbf{D}^{(k)})}^{-1} \S^{(k)} \mathbf{D}^{(k)} =  {(b^{(k)})}^{-1}  \circ \S^{(k)} \circ {(b^{(k)})}^{T}.
\end{equation}

Eq.~\eqref{eq: mulskd} gives an explicit way to compute $\uppdelta^{(k)}$, which only requires scanning the non-zero elements in $\S^{(k)}$. Let $s$ be the number of non-zero elements in matrix $\S$.
Then, the time cost to obtain $\uppdelta^{(k)}$ is $O(ks)$. Since $k$ is a small number and $\S$ is often sparse for DAG, the time cost $O(ks)$ is close to $O(d)$. Meanwhile, the space cost to obtain $\uppdelta^{(k)}$ is at most $O(s)$. Therefore, it is time and space efficient to compute $\delta$.

 \smallskip
\noindent \underline{\textbf{Computing $\grd_{\S} \uppdelta^{(k)}$.}}
We now turn our attention to computing $\grd_{\S} \uppdelta^{(k)}$, which clearly exists
according to Eq.~\eqref{eq: uppdelta} so R2 is satisfied. We will now manually apply backward differentiation to the iteration defining $\grd_{\S} \uppdelta^{(k)}$ (sometimes referred to as the ``adjoint'' method outside of the ML literature). As a result, we obtain a recipe for iteratively computing $\grd_{\S} \uppdelta^{(k)}$. This iterative method will allow us to implement the gradient such that $\grd_{\S} \uppdelta^{(k)}$ remains \emph{sparse} throughout the entire process.

To simplify notation, we denote the Hadamard product of two vectors $u^{\alpha} \circ v^{-\alpha} $ as $ {(\frac{u}{v})}^{\alpha}$. By Eq.~\eqref{eq: mulskd} and Lemma~\ref{lem: uppdelta}, following the chain rule, we obtain explicit formulae for $\grd_{\S} \uppdelta^{(k)}$ as follows.

\begin{lemma}
\label{lem: grdD2S}
For any non-negative matrix $\S$, $k \geq 0$ and $0 \leq \alpha \leq 1$, we always have
\begin{equation}
\grd_{\S^{(k)}} \uppdelta^{(k)}  = \grd_{\S^{(k)}} b^{(k)}  = x^{(k)} \circ \mathbf{J} + {(y^{(k)})}^{T} \circ \mathbf{J},
\end{equation}
where $x^{(k)} = \alpha { \left(\frac{{c (\S^{(k)})}}{r(\S^{(k)})} \right)}^{1 - \alpha} $,
$y^{(k)} = (1 - \alpha) { \left(\frac{{r(\S^{(k)})}}{{{c(\S^{(k)})}}} \right)}^{\alpha}$, and $\mathbf{J} \in \mathbb{R}^{d \times d}$ is a matrix with all entries equal to $1$.
\end{lemma}

\begin{lemma}
\label{lem: grdSj2Sj1}
For any non-negative matrix $\S$, $k \geq 1$ and $0 \leq \alpha \leq 1$, 
given any $1 \leq j \leq k$, suppose that we already have $\grd_{\S^{(j)}} \uppdelta^{(k)}$, let
\begin{equation}
\label{eq: grdzj}
\begin{split}
z^{(j - 1)} & = - \frac{r(\grd_{\S^{(j)}} \uppdelta^{(k)} \circ \S^{(j-1)} \circ {(b^{(j-1)})}^{T} )}{{(b^{(j - 1)})}^{2}} \\
& + c( {(b^{(j - 1)})}^{-1} \circ \grd_{\S^{(j)}} \uppdelta^{(k)} \circ \S^{(j - 1)}).
\end{split}
\end{equation}
Then, we have
\begin{equation}
\begin{split}
~ & \grd_{\S^{(j - 1)}} \uppdelta^{(k)}   =  {(b^{(j - 1)})}^{-1} \circ \grd_{\S^{(j)}} \uppdelta^{(k)} \circ 
{(b^{(j - 1)})}^{T} \\
& + x^{(j - 1)} \circ z^{(j - 1)} \circ \mathbf{J} + {(y^{(j - 1)})}^{T} \circ {(z^{(j - 1)})}^{T} \circ \mathbf{J},
\end{split}
\end{equation}
where $x^{(j - 1)}$ and $y^{(j - 1)}$ have the same meaning as Lemma~\ref{lem: grdD2S}.
\end{lemma}

Now, if we directly compute $\grd_{\S} \uppdelta^{(k)}$ following the above lemmas, the resulting algorithm would not be space efficient since $\grd_{\S^{(k)}} \uppdelta^{(j)}$ would be a dense matrix for all $0 \leq j \leq k$. Nevertheless, since the final objective is to compute $\grd_{\W} \uppdelta^{(k)} = 2 \grd_{\S} \uppdelta^{(k)} \circ \W$, it is only necessary to compute the gradient of non-zero elements in $\W$ and $\S$. Interestingly, we find that it is also safe to do this ``masking'' in advance. The correctness is guaranteed by the following lemma.

\begin{lemma}
\label{lem: spvalidgrd}
For any matrix $\W$, $k \geq 1$ and $0 \leq \alpha \leq 1$, let $\M \in \mathbb{R}^{d \times d}$ be such that $\M[i, j] = 1$ when $\W[i, j] \neq 0$ and $\M[i, j] = 0$ otherwise. Let
$\grd'_{\S^{(k)}} \uppdelta^{(k)}  = x^{(k)} \circ \M  + {(y^{(k)})}^{T} \circ \M$.
For all $1 \leq j \leq k - 1$, let
\vspace{-0.5em}
\begin{equation}
\label{eq: MgrdSj2Sj1}
\begin{split}
~ & \grd'_{\S^{(j - 1)}} \uppdelta^{(k)}  =  {(b^{(j - 1)})}^{-1} \circ \grd'_{\S^{(j)}} \uppdelta^{(k)} \circ 
{(b^{(j - 1)})}^{T} \\
+ & ~ {x}^{(j - 1)} \circ z^{(j - 1)} \circ \mathbf{M} + {(y^{(j - 1)})}^{T} \circ {(z^{(j - 1)})}^{T} \circ \mathbf{M}, 
\end{split}
\end{equation}
where $x^{(j - 1)}$ and $y^{(j - 1)}$ are defined in Lemma~\ref{lem: grdD2S},
and $z^{(j - 1)}$ is defined in Eq.~\eqref{eq: grdzj}.
Then, we always have 
\begin{equation}
\grd_{\W} \uppdelta^{(k)} = 2 \grd'_{\S} \uppdelta^{(k)} \circ \W.
\end{equation}
\end{lemma}

According to Lemma~\ref{lem: spvalidgrd}, we can obtain $\grd_{\W} \uppdelta^{(k)}$ in a space efficient way by properly making use of the sparsity structure of $\W$ and $\S$. For each $1 \leq j \leq k - 1$, the time cost to compute $x^{(j)}$, $y^{(j)}$ and $h^{(j)}$ are all $O(s)$. Therefore, the time cost to compute $\grd_{\W} \uppdelta^{(k)}$ is also $O(ks)$, which is close to $O(d)$ for sparse $\W$.

\smallskip
\noindent \underline{\textbf{Procedures Description.}}
To summarize, we describe the whole process to compute $\uppdelta^{(k)}$ and $\grd_{\W} \uppdelta^{(k)}$ in Fig.~\ref{fig: algcompute}. Note that computing both $\uppdelta^{(k)}$ and $\grd_{\W} \uppdelta^{(k)}$ will cost at most $O(ks)$ time and $O(s)$ space, near linear $O(d)$ cost for DAGs.
Therefore, the requirements R3 and R4 are satisfied.

%According to our experimental results presented in Section~\ref{sec: exp}, using this new acyclicity constraint enables our structure learning algorithm, in some cases, to run $1$--$2$ orders of magnitude faster than \textsf{NOTEARS} and scales to (sparse) graphs with possibly millions of nodes.

\begin{figure}[t]
\small
\rule{\linewidth}{1pt}
\leftline{\textbf{Procedure} \textsc{Forward$(\W, k, \alpha)$}}
\vspace{-1em}
	\begin{algorithmic}[1]
		\STATE $\S^{0} = \W^{2}$
		\FOR{$j = 0 \text{ to } k$}
		\STATE $b^{(j)} \gets {(r(\S^{(j)}))}^{\alpha} \circ {(c(\S^{(j)}))}^{1 - \alpha}$  
		\IF{$j \leq k - 1$}
		\STATE compute $\S^{(j + 1)}$ by Eq.~\eqref{eq: mulskd}
		\ENDIF
		\ENDFOR
		\RETURN $\uppdelta^{(k)} \gets \sum_{i = 1}^{d} b^{(k)}[i]$
	\end{algorithmic}
\rule{\linewidth}{0.5pt}\\
\leftline{\textbf{Procedure} \textsc{Backward$(\W, k, \alpha)$}}
\vspace{-1em}
	\begin{algorithmic}[1]
		\STATE $\M \gets \W^{0}$
		\STATE compute $x^{(k)}$ and $y^{(k)}$ by Lemma~\ref{lem: grdD2S}
		\STATE compute $\grd'_{\S^{(k)}} \uppdelta^{(k)}$ by Lemma~\ref{lem: spvalidgrd}
		\FOR{$j = k \text{ to } 1$}
		\STATE compute $x^{(j - 1)}$ and $y^{(j - 1)}$ by Lemma~\ref{lem: grdD2S}
		\STATE compute $z^{(j - 1)}$ by Eq.~\eqref{eq: grdzj}
		\STATE compute $\grd'_{\S^{(j - 1)}} \uppdelta^{(k)}$ by Eq.~\eqref{eq: MgrdSj2Sj1}
		\ENDFOR
		\RETURN $\grd_{\W} \uppdelta^{(k)} \gets 2 \grd'_{\S^{(0)}} \uppdelta^{(k)} \circ \W$
	\end{algorithmic}
	\rule{\linewidth}{1pt}
		\vspace{-1em}
	\caption{\normalsize Procedures for computing $\uppdelta^{(k)}$ and $\grd_{\W} \uppdelta^{(k)}$.}
	\label{fig: algcompute}
	\vspace{-1em}
\end{figure}

%% file: algorithm.tex
% !TeX spellcheck = en_US

\section{Structure Learning Algorithm}
\label{sec: com}

In this section, we propose the structure learning algorithm built on top of the new acyclicity constraint introduced in the previous section. Our algorithm is called \textsc{LEAST}, which represents a \underline{L}arge-scale, \underline{E}fficient and \underline{A}ccurate \underline{S}truc\underline{T}ure learning method for BNs. First, we present the details of the algorithm in our exemplary LSEM case. Then, we shortly discuss its implementation details.

\begin{figure}[t]
	\small
	\rule{\linewidth}{1pt}
	\leftline{\textbf{Algorithm} \textsc{LEAST$(X, \zeta, \lambda, \epsilon, k, \alpha, B, \theta, T_{o}, T_{i})$}}
	\vspace{-1em}
	\begin{algorithmic}[1]
		\STATE $\rho \gets 1$, $\eta \gets 1$
		\REPEAT
		\STATE $\W^{*}, \uppdelta(\W^{*}) \gets \textsc{Inner}(X, \zeta, \lambda, \rho, \eta, k, \alpha, B, \theta, T_{i})$
		\STATE $\eta \gets \eta + \rho \uppdelta(\W^{*})$
		\STATE enlarge $p$ by a small factor
		\UNTIL{$\uppdelta(\W^{*}) \leq \epsilon$ or running $T_o$ iterations}
		\RETURN $\W^{*}$
	\end{algorithmic}
	\vspace{-0.5em}
	\rule{\linewidth}{0.5pt}\\
	\leftline{\textbf{Procedure} \textsc{Inner$(X, \zeta, \lambda, \rho, \eta, k, \alpha, \theta, T)$}}
	\vspace{-1em}
	\begin{algorithmic}[1]
		\STATE randomly initialize $\W$ as a sparse matrix with density $\zeta$ using Glorot uniform initialization
		\REPEAT
		\STATE $\uppdelta(\W) \gets \textsc{Forward}(\W, k, \alpha)$
		\STATE $\grd_{\W} \uppdelta(\W) \gets \textsc{Backward}(\W, k, \alpha)$
		\STATE randomly fetch a sample $\X_{B}$ from $X$ with $B$ samples
		\STATE $\ell(\W) \gets L(\W, \X_{B}) + \frac{\rho}{2}  {\uppdelta(\W)}^{2}+ \eta \uppdelta(\W)$
		\STATE $\grd_{\W} \ell(\W) \gets \grd_{\W} L(\W, \X_{B}) + (\rho + \uppdelta(\W)) \grd_{\W} \uppdelta(\W) $
		\STATE update $\W$ with some optimizer using $\grd_{\W} \ell(\W)$ $\%$ e.g. Adam
		\STATE filter all elements in $\W$ whose absolute value is less than $\theta$
		\UNTIL{$\ell(\W)$ converges or running $T$ iterations}
		\RETURN $\W$ and $\uppdelta(\W)$
	\end{algorithmic}
	\rule{\linewidth}{1pt}
	\vspace{-1em}
	\caption{\normalsize Algorithm for large-scale BN structure learning.}
	\vspace{-1em}
	\label{fig: algsuperBN}
\end{figure}

\noindent \underline{\textbf{\algo Algorithm.}}
Given a matrix $\mathbf{A}$, let ${\| \mathbf{A} \|}_{p}$ denote the $p$-norm of $\mathbf{A}$ for all $p \geq 1$. Following~\cite{zheng2018dags}, we set the loss function in Eq.~\eqref{eq: slp-lsem} as $L(\W, \X) = \frac{1}{n} {\| \X - \X \W \|}_{2}^{2}  + \lambda {\| \W \|}_{1}$, that is the least squares with $L_{1}$-regularization. We denote  $\uppdelta(\W)$ as the spectral radius upper bound of the matrix $\W$. To solve this optimization problem, we use the augmented Lagrangian~\cite{lukasik2009firefly} method with some adjustments. 
Concretely, we solve the following program augmented from Eq.~\eqref{eq: slp-lsem} with a quadratic penalty: 
\begin{equation}
\label{eq: slp-lsem-al}
\begin{split}
\arg\min\nolimits_{\W} \frac{1}{n} {\| \X - \X \W \|}_{2}^{2}  & + \lambda {\| \W \|}_{1} + \frac{\rho}{2} {\uppdelta(\W)}^{2}, \\
\text{ subject to }  \uppdelta & (\W)  = 0,
\end{split}
\end{equation}
where $\rho > 0$ is a penalty parameter. Eq.~\eqref{eq: slp-lsem-al} is then transformed into an unconstrained program. Specifically, let $\eta$ denote the Lagrangian multiplier, and $\ell(\W) = L(\W, \X) + \frac{\rho}{2}  {\uppdelta(\W)}^{2}+ \eta \uppdelta(\W)$ represent the unconstrained objective function. Solving Eq.~\eqref{eq: slp-lsem-al} then consists  of two iterative steps:

1) find the optimal value $\W^{*}$ minimizing the function $\ell(\W)$;

2) increase the multiplier $\eta$ to $\eta + \rho \uppdelta(\W)$ and optionally increase the penalty value $\rho$ to improve convergence.

With the growth of $\rho$ and $\eta$, since $\uppdelta(\W) \geq 0$, minimizing $\ell(\W)$ forces $\uppdelta(\W)$ to decrease to near $0$. \algo follows this generic process. In each iteration, we use the \textsf{Inner} procedure to optimize $\ell(\W)$ using a first gradient based method (line~3), such as Adam~\cite{kingma2014adam}, and then enlarge $\rho$ and $\eta$ (lines~4--5). Finally, we return $\W$ when $\uppdelta(\W)$ is below a small tolerance value $\epsilon$. In each iteration, we initialize $\W$ as a random sparse matrix with density $\zeta$ using Glorot uniform initialization (line~1). 
To improve efficiency and scalability, we also incorporate the two techniques into \textsc{LEAST}:

1) \textbf{Batching:} in line~5 of \textsc{Inner}, we randomly fetch a batch $\X_B$ from $\X$ to optimize $\ell(\W)$ instead of using $\X$. 
% Using batches can improve the computational efficiency for $B << n$ but does not appear to impact final results' accuracy. This is because gradient descent algorithms on stochastic batches could also converge, and has been adopted as a standard method in machine learning field~\cite{bottou2012stochastic, kingma2014adam}.

2) \textbf{Thresholding:} in line~9 of the \textsf{Inner} procedure, we filter elements in $\W$ with absolute value below a small threshold $\theta$. A small numerical value in $\W$ indicates a weak correlation between the two nodes, so filtering out them may help ruling out false cycle-inducing edges in advance. It has been shown that this helps to decrease the number of false positive edges~\cite{wang2016no, zheng2018dags, zhou2009thresholding}. Moreover, removing these elements makes $\W$ remain sparse throughout the optimization process, thus ensuring overall computational efficiency.

\smallskip
\noindent \underline{\textbf{Complexity Analysis.}}
We now analyze the computational complexity of each round in \textsc{LEAST}. 
Computing $L(\W,\ X_{B})$ and its gradient $\grd_{\W} L(\W,\ X_{B})$ costs $O(Bsd)$ time and $O(s + Bd)$ space, where $s$ is the number of non-zero elements in $\W$. As a result, the time cost for computing the acyclicity constraint $O(s) << O(Bsd)$. Thus, overall training time will be dominated by the first term in Eq.~\eqref{eq: slp-lsem-al}. 
 Meanwhile, since $\W$ remains sparse and $\X_{B}$ is a small matrix, they can easily fit into memory.

\smallskip
\noindent \underline{\textbf{Implementation Details.}}
We implement two versions of our \algo algorithm in Python:

1) \textsc{LEAST-TF} is a Tensorflow-based implementation using dense tensors only. We rely on Tensorflow's automatic differentiation capabilities to compute gradients.

2) \textsc{LEAST-SP} based on the SciPy library, where we have implemented \algo using sparse data structures. We use the CSR format to represent the sparse matrices. Gradients for the constraint function $\uppdelta$ are implemented according to the description in Section \ref{sec: nco}. We use the Adam~\cite{kingma2014adam} optimizer, since it  exhibits fast convergence and does not generate dense matrices during the computation process.

The two implementations are used in different scenarios. 
We generally use \textsc{LEAST-TF} for those cases where a dense $\W$ may fit entirely into the (GPU) memory. This is particularly relevant for those applications which have strict requirements on training speed, such as real-time monitoring systems presented in Section~VI-A. We also use \textsc{LEAST-TF} to validate our method with respect to our accuracy and efficiency claims in the following Section~V.

For use cases where data contains hundreds of thousands variables and a high degree of sparsity, we use \textsc{LEAST-SP}. 
By leveraging sparse data structures, \textsc{LEAST-SP} is able to deal with cases where the dense representation $\W$ does not fit into GPU or main memory anymore. It is thus also used for our scalability testing experiments in the next section.

We have deployed both \textsc{LEAST-TF} and \textsc{LEAST-SP}
in the production environment of our company. They are part of an internal library supporting different applications in Alibaba. 
Until now, they have been used in more than 20 business scenarios including production system monitoring, 
recommendation systems, health care data analysis, money
laundering detection, could security and etc. It is executed thousands of times per day. We also plan to release an open-source implementation of \textsc{LEAST} to the research community.

%% file: exp.tex
% !TeX spellcheck = en_US

\section{Evaluation Results}
\label{sec: exp}

We conduct extensive experiments to evaluate the accuracy, efficiency and scalability of our proposed method. The results are reported in this section.

\smallskip
\noindent \underline{\textbf{Algorithms.}}
We compare our \textsc{LEAST} algorithm with \textsc{NOTEARS},
the state-of-the-art structure learning algorithm for BN proposed in~\cite{zheng2018dags}. It provides the most straightforward way to establish the correctness of our proposed method, while allowing for a fair comparison w.r.t.~computational efficiency. 
We refrain from comparing with traditional combinatorial optimization algorithms such as~\cite{ramsey2017million, shimizu2011directlingam, spirtes2000causation}. The reason is that we obtain results consistent with~\cite{zheng2018dags}, which already shows that \textsc{NOTEARS} generally outperforms them.

\smallskip
\noindent \underline{\textbf{Parameter Settings.}}
In our \algo method, for the upper bound $\uppdelta(\W)$, we set $k = 5$ and the balancing factor $\alpha = 0.9$. For the optimizer in the \textsc{Inner} procedure, we use Adam~\cite{kingma2014adam} and set its learning rate to $0.01$. We set the initialization density $\zeta = 10^{-4}$. We furthermore ensure that in \textsc{LEAST-SP} Adam is operating on sparse matrices only. The maximum outer and inner iteration numbers are set to $1,000$ and $200$ for all algorithms, respectively. Remaining parameters are tuned individually for each experiment.

For \textsc{NOTEARS} we use the Tensorflow implementation provided in~\cite{lee2019scaling}. Note that, it seems hardly possible to implement \textsc{NOTEARS} purely using sparse matrices, as some steps would always involve completely dense matrices.

\smallskip
\noindent \underline{\textbf{Environment.}}
All of our experiments are run on a Linux server with a Intel Xeon Platinum 2.5 GHz CPU with 96 cores and 512GB DDR4 main memory. For experiments on GPUs  we use a  NVIDIA Tesla V100 SXM2 GPU with 64GB GPU memory. 
We run each experiment multiple times and report the mean value of each evaluation metric.

In the following, Section~\ref{sec: exp-art} reports the evaluation results on artificial benchmark datasets in terms of accuracy and efficiency. Section~\ref{sec: exp-rel} examines the scalability of our method on large-scale real-world datasets.

\input{figs.tex}

\subsection{Artificial Datasets}
\label{sec: exp-art}

We use the graph generation code from~\cite{zheng2018dags} to produce benchmark artificial data. 
It generates a random graph topology of $G$ following two models, Erdös-Rényi (ER) or scale-free (SF), and then assigns each edge a uniformly random weight to obtain the adjacency matrix $\W$. The sample matrix $\X$ is then generated according to LSEM with three kinds of additive noise: Gaussian (GS), Exponential (EX), and Gumbel (GB). We vary the node size $d$, the sample size $n$ and the average degree of nodes.  Following~\cite{zheng2018dags}, we set  the average node degree to $2$ for ER and $4$ for SF graphs.
For fairness, we compare \textsc{LEAST-TF}, the Tensorflow implementation of \textsc{LEAST} with \textsc{NOTEARS}.
We also slightly modify the termination condition of \textsc{LEAST}.
At the end of each outer loop, we also compute the value of $h(\W)$ and terminate when $h(\W)$ is smaller than the tolerance value $\epsilon$. In this way, we ensure that their convergence  is tested using the same termination condition. For remaining parameters, we set the batch size $B$ equal to the sample size $n$, the filtering threshold $\theta = 0$ and the regularization penalty factor $\lambda = 0.5$.
All experiments in this subsection are run on the CPU.

\input{figs-sca.tex}

\smallskip
\noindent \underline{\textbf{Result Accuracy.}}
We evaluate by comparing results with the original ground truth graph $G$ used to generate random samples.
Following~\cite{zheng2018dags}, after optimizing the result matrix $\W$ to a small tolerance value $\epsilon$, we filter it using a small threshold $\tau$ to obtain $\W'$, and then compare $G(\W')$ with $G$. 
We apply a grid search for the two hyper-parameters $\epsilon \in \{10^{-1}, 10^{-2}, 10^{-3}, {10^{-4}}\}$ and $\tau \in \{0.1, 0.2, 0.3, 0.4, 0.5\}$, and report the result of the best case. We vary $d = {10, 20, 50, 100}$ and set $n = 10d$.
The results for the two graph models with three types of noise are shown in Fig.~\ref{fig: artiexp-acc}. 

The first two lines of Fig.~\ref{fig: artiexp-acc} report the results in terms of $F_1$-score and the Structural Hamming Distance (SHD), respectively. The third line reports the detailed Pearson correlation coefficients of $\uppdelta(\W)$ and $h(\W)$ recorded during the computation process of \algo method.
The error bar indicates the standard deviation of each value. We have the following observations:

1) For \algo we obtain $F_1$-scores which are larger than $0.8$ in almost all cases. Meanwhile, the results for \algo are very close to those of \textsc{NOTEARS} in terms of the $F_1$-score and SHD. The difference appears to be negligible in most cases.

2) Our upper bound based acyclicity measure $\uppdelta(\W)$ is highly correlated with  the original metric $h(\W)$, with correlation coefficients larger than $0.8$ in all cases and larger than $0.9$ in most cases. This verification indicates that $\uppdelta(\W)$ is consistent with $h(\W)$ and a valid acyclicity measure. 

3) Results of \algo appear to have higher variance than those of  \textsc{NOTEARS}. This is likely due to small numerical instabilities in the iterative computation process for  $\uppdelta(\W)$.

4) the observed difference in variance between \algo and \textsc{NOTEARS} is larger on dense SF-4 graphs than sparse ER-2 graphs, also more noticeable for large $d \geq 50$ than small $d$. Interestingly, as soon as the variance increases,  correlation coefficients decrease at the same time. This could be attributed to the number of non-zero elements in $\W$, which increases w.r.t.~$d$ and the graph density.

Overall, this set of experiments shows that, for the above graphs, our proposed acyclicity metric $\uppdelta(\W)$ produces results consistent with the original metric $h(\W)$. Yet, we note that the increased computational efficiency (see the following results) comes at the price of an increased variance.

\smallskip 
\noindent \underline{\textbf{Time Efficiency.}}
The evaluation results are reported in the forth row of Fig.~\ref{fig: artiexp-acc}, which represents the execution time by fixing $\epsilon = 10^{-4}$ and sampling size $n = 10d$. We find that:

1) \algo runs faster than  \textsc{NOTEARS} in all tested cases.
The speed up ratio ranges from $5$ to $15$. This is mainly due to the fact that the time complexity of computing the acyclicity constraint in the case of \algo is much lower than that in \textsc{NOTEARS} (near $O(d)$ vs. $O(d^3)$). 

2) the speed up effects of \algo are similar for different graph models and noise types. This is because the time complexity of \algo only depends on $d$, but is independent of all other factors.

3) the speed up effect is more obvious when $d$ gets larger. The speed up ratio is up to $10$, $9.5$ and $14.7$ when $d = 100$, $200$ and $500$, respectively. This is because the time cost for evaluating the acyclicity constraint grows almost linearly with $d$ in \algo while cubic with $d$ in \textsc{NOTEARS}.

Overall, this set of experiments indicates that \algo generally outperforms \textsc{NOTEARS} in terms of computational efficiency, especially for larger graphs.

\smallskip 
\noindent \underline{\textbf{Summary.}}
On artificial benchmark data, our \algo algorithm attains comparable accuracy w.r.t.~the state-of-the-art \textsc{NOTEARS} method while improves the time efficiency by $1$--$2$ orders of magnitude.

\subsection{Real-World Datasets}
\label{sec: exp-rel}

We examine the scalability of our method on  large-scale real-world datasets. Their properties of are summarized in Table~\ref{Tab: Dataset}. It should be noted that it is difficult to find suitable, publicly available benchmark datasets which involve more than thousands of variables. This is why we also present some results based on non-public, internal datasets.
\begin{itemize}[leftmargin = *]
	\item \textsl{Movielens} is based on the well-known MovieLens20M~\cite{harper2015movielens} recommendation dataset. We regard each movie as a node and each user's entire rating history as one sample. The rating values range from $0$ to $5$ with an average rating of about $3.5$. We construct the data matrix $\X$ in a standard way as follows: Let $r_{ij}$ be the rating of user $i$ for movie $j$. Then, we set $\X[i, j] = r_{ij} - \sum_{j=1}^{n_i} r_{ij}/n_i$, where $n_i$ is the number of available ratings for user $i$, so that we subtract each user's mean rating from their ratings. We present a case study using this dataset later in Section~VI-C.
	
	\item \textsl{App-Security} and \textsl{App-Recom} are two datasets extracted from real application scenarios at Alibaba. They are used in the context of cloud security and recommendation systems, respectively. Due to the sensitive nature of these businesses, we cannot provide more details beyond their size.
\end{itemize}

\begin{table}[t]
	\centering
	\caption{Properties of real-world large-scale datasets.}

		\begin{tabular}{ccc}
			\hline
			\textbf{Dataset Name} & \textbf{\# of Nodes} & \textbf{\# of Samples} \\ \hline
			\textsl{Movielens} & $27,278$ & $138,493$ \\ \hline
			\textsl{App-Security} &$91, 850$ & $1,000, 000$  \\ 
			\textsl{App-Recom} &$159,008$ & $584,871$  \\ 
			\hline
	\end{tabular}
	\label{Tab: GeneRes}
		\vspace{-1.5em}
\end{table}

We set the batch size $B = 1, 000$, the filtering threshold $\theta = 10^{-3}$ and the tolerance value $\epsilon = 10^{-8}$ and run \textsc{LEAST-SP}, the sparse matrices based implementation of \textsc{LEAST} on them. Note that, \textsc{NOTEARS} is unable to scale to these datasets. Fig.~\ref{fig: artiexp-sca} illustrates how the constraint value $\uppdelta(\W)$ and $h(\W)$ varies with regard to the execution time. We find that:

1) In all cases, when optimizing the constraint $\uppdelta(\W)$, the value of $h(\W)$ also decreases accordingly and converges to a very small level. This verifies the effectiveness of our proposed acyclicity constraint on large graphs.

2) Our proposed algorithm can scale to very large graphs while taking a reasonable time. It takes $89.4$, $67.2$ and $6.5$ hours on the datasets \textsl{Movielens}, \textsl{App-Security} and \textsl{App-Recom}, respectively. 
So far, to the best of our knowledge, no existing continuous optimization structure learning algorithm can process SLP with more than $10^{4}$ nodes, whereas our proposed method successfully scales to at least $10^5$ nodes.

%% file: figs.tex
\pgfplotsset{width = 4.0cm}
\pgfplotsset{compat = 1.15}

% Figure for accuracy evaluation results on F1 value
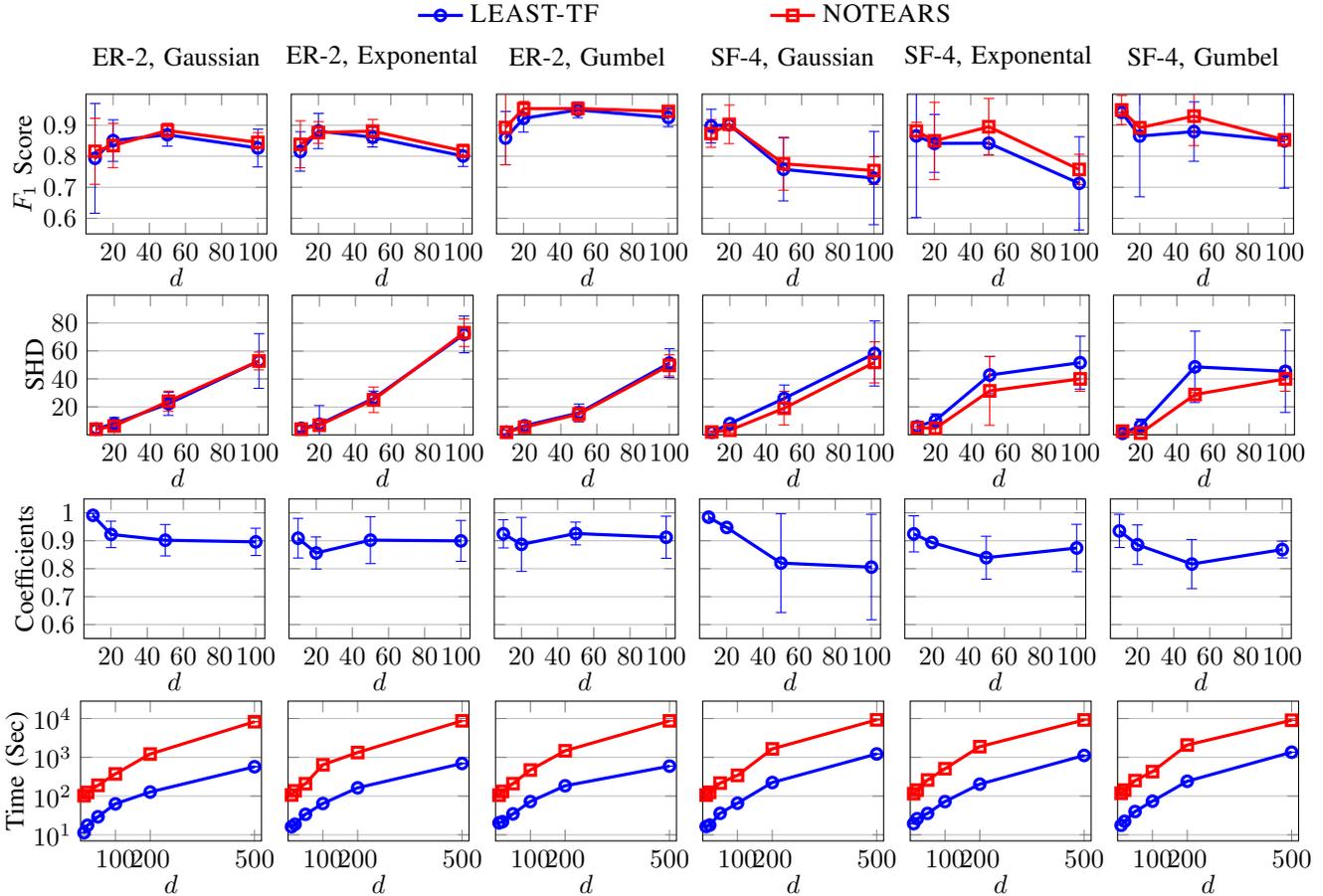
\begin{figure*}
    \hspace{15em} \ref{artileg}
    
	\begin{tikzpicture} % ER-2, Gaussian
	\begin{axis}[
	title = {ER-2, Gaussian},
	xlabel = {$d$}, xmin = 5, xmax = 105, xtick = {20, 40, 60, 80, 100}, xlabel shift = -4, 
	ylabel = {$F_1$ Score}, ymin = 0.55, ymax = 1, ytick = {0.6, 0.7, 0.8, 0.9}, ylabel shift = -4, ymajorgrids=true, 
	legend style = {draw= none},
	legend columns=-1,
    legend entries={ \textsc{LEAST-TF \quad \quad \quad \quad \quad \quad \quad},\textsc{NOTEARS}},
    legend to name = artileg
	]
	\addplot+ [
	mark = o, very thick,
	error bars/.cd,
	y dir= both,y explicit,
	] coordinates {
		(10,0.7933)  +- (0.1769, 0.1769) %
		(20,0.8503)  +- (0.0670, 0.0670) %
		(50,0.8691)  +- (0.0368, 0.0368) %
		(100,0.8264)  +- (0.0610, 0.0610)%
	};
	
	\addplot+ [
	mark=square, very thick, 
	error bars/.cd,
	y dir= both,y explicit
	] coordinates {
		(10,0.8158)  +- (0.1062, 0.1062) %
		(20,0.8345)  +- (0.0711, 0.0711) % 
		(50,0.8827)  +- (0.0235, 0.0235) %
		(100,0.8447)  +- (0.0315, 0.031) %
	};
	\end{axis}
	\end{tikzpicture}
	\hspace{-1.2em}
	\begin{tikzpicture} % ER-2, Exponental
	\begin{axis}[
	title = {ER-2, Exponental},
	xlabel = {$d$}, xmin = 5, xmax = 105, xtick = {20, 40, 60, 80, 100}, xlabel shift = -4, 
	ymin = 0.55, ymax = 1, ytick = {0.6, 0.7, 0.8, 0.9}, yticklabel = \empty, ymajorgrids=true
	]
	\addplot+ [
	mark = o, very thick, 
	error bars/.cd,
	y dir= both,y explicit,
	] coordinates {
		(10,0.8153)  +- (0.0636, 0.0636) %
		(20,0.8812)  +- (0.0568, 0.0568) %
		(50,0.8610)  +- (0.0313, 0.0313) %
		(100,0.8006)  +- (0.0337, 0.0337) %
	};
	
	\addplot+ [
	mark = square, very thick, 
	error bars/.cd,
	y dir= both,y explicit,
	] coordinates {
		(10,0.8384)  +- (0.0755, 0.0755) %
		(20,0.8762)  +- (0.0350, 0.0350) %
		(50,0.8807)  +- (0.0377, 0.0377) %
		(100,0.8171)  +- (0.0211, 0.0211) %
	};
	\end{axis}
	\end{tikzpicture}
	\hspace{-1.2em}
	\begin{tikzpicture} % ER-2, Gumbel
	\begin{axis}[
	title = {ER-2, Gumbel},
	xlabel = {$d$}, xmin = 5, xmax = 105, xtick = {20, 40, 60, 80, 100}, xlabel shift = -4, 
	ymin = 0.55, ymax = 1.0, ytick = {0.6, 0.7, 0.8, 0.9}, yticklabel = \empty, ymajorgrids=true
	]
	\addplot+ [
	mark = o, very thick, 
	error bars/.cd,
	y dir= both,y explicit,
	] coordinates {
		(10,0.8583)  +- (0.0858, 0.0858) %
		(20,0.9221)  +- (0.0441, 0.0441) %
		(50,0.9491)  +- (0.0253, 0.0253) %
		(100,0.9245)  +- (0.0297, 0.0297) %
	};
	
	\addplot+ [
	mark = square, very thick, 
	error bars/.cd,
	y dir= both,y explicit,
	] coordinates {
		(10,0.8928)  +- (0.1207, 0.1207) %
		(20,0.9532)  +- (0.0238, 0.0238) %
		(50,0.9536)  +- (0.0146, 0.0146) %
		(100,0.9445)  +- (0.0051, 0.0051) %
	};
	\end{axis}
	\end{tikzpicture}
	\hspace{-1.2em}
	\begin{tikzpicture} % SF-4, Gaussian
	\begin{axis}[
	title = {SF-4, Gaussian},
	xlabel = {$d$}, xmin = 5, xmax = 105, xtick = {20, 40, 60, 80, 100}, xlabel shift = -4, 
	ymin = 0.55, ymax = 1.0, ytick = {0.6, 0.7, 0.8, 0.9}, yticklabel = \empty, ymajorgrids=true
	]
	\addplot+ [
	mark = o, very thick, 
	error bars/.cd,
	y dir= both,y explicit,
	] coordinates {
		(10,0.8972)  +- (0.0544, 0.0544) %
		(20,0.9019)  +- (0, 0) %
		(50,0.7577)  +- (0.1017, 0.1017) %
		(100,0.7295)  +- (0.15, 0.15) %
	};
	
	\addplot+ [
	mark = square, very thick, 
	error bars/.cd,
	y dir= both,y explicit,
	] coordinates {
		(10,0.8732)  +- (0.0447, 0.0447) %
		(20,0.9028)  +- (0.0622, 0.0622) %
		(50,0.7758)  +- (0.0853, 0.0853) %
		(100,0.7537)  +- (0.0446, 0.0446) %
	};
	\end{axis}
	\end{tikzpicture}
	\hspace{-1.2em}
	\begin{tikzpicture} % SF-4, Exponental
	\begin{axis}[
	title = {SF-4, Exponental},
	xlabel = {$d$}, xmin = 5, xmax = 105, xtick = {20, 40, 60, 80, 100}, xlabel shift = -4, 
	ymin = 0.55, ymax = 1.0, ytick = {0.6, 0.7, 0.8, 0.9}, yticklabel = \empty, ymajorgrids=true
	]
	\addplot+ [
	mark = o, very thick, 
	error bars/.cd,
	y dir= both,y explicit,
	] coordinates {
		(10,0.8655)  +- (0.2629, 0.2629) %
		(20,0.8412)  +- (0.0932, 0.0932) %
		(50,0.8419)  +- (0.0372, 0.0372) %
		(100,0.7126)  +- (0.1501, 0.1501) %
	};
		
	\addplot+ [
	mark = square, very thick, 
	error bars/.cd,
	y dir= both,y explicit,
	] coordinates {
		(10,0.8811)  +- (0.0280, 0.0280) %
		(20,0.8495)  +- (0.1244, 0.1244) %
		(50,0.8949)  +- (0.0916, 0.0916) %
		(100,0.7574)  +- (0.0487, 0.0487) %
	};
	\end{axis}
	\end{tikzpicture}
	\hspace{-1.2em}
	\begin{tikzpicture} % SF-4, Gumbel
	\begin{axis}[
	title = {SF-4, Gumbel},
	xlabel = {$d$}, xmin = 5, xmax = 105, xtick = {20, 40, 60, 80, 100}, xlabel shift = -4, 
	ymin = 0.55, ymax = 1.0, ytick = {0.6, 0.7, 0.8, 0.9}, yticklabel = \empty, ymajorgrids=true
	]
	\addplot+ [
	mark = o, very thick, 
	error bars/.cd,
	y dir= both,y explicit,
	] coordinates {
		(10,0.9412)  +- (0, 0) %
		(20,0.8649)  +- (0.1956, 0.1956) %
		(50,0.8793)  +- (0.0957, 0.0957) %
		(100,0.8493)  +- (0.1523, 0.1523) %
	};
		
	\addplot+ [
	mark = square, very thick, 
	error bars/.cd,
	y dir= both,y explicit,
	] coordinates {
		(10,0.9490)  +- (0.0472, 0.0472) %
		(20,0.8918)  +- (0.0164, 0.0164) %
		(50,0.9291)  +- (0.0950, 0.0950) %
		(100,0.8529)  +- (0.0021, 0.0021)
	};
	\end{axis}
	\end{tikzpicture}

% Figure for accuracy evaluation results on SHD
	\hspace{0.1em}
	\begin{tikzpicture} % ER-2, Gaussian
	\begin{axis}[
	xlabel = {$d$}, xmin = 5, xmax = 105, xtick = {20, 40, 60, 80, 100}, xlabel shift = -4, 
	ylabel = {SHD}, ymin = 0, ymax = 100, ytick = {20, 40, 60, 80}, ylabel shift = -4, ymajorgrids=true
	]
	\addplot+ [
	mark = o, very thick, 
	error bars/.cd,
	y dir= both,y explicit,
	] coordinates {
		(10,4.2)  +- (3.3106, 3.3106) %
		(20,8)  +- (4.4272, 4.4272) %
		(50, 22.2)  +- (8.3283, 8.3283) %
		(100,52.8)  +- (19.5694, 19.5694) %
	};
	
	\addplot+ [
	mark = square, very thick, 
	error bars/.cd,
	y dir= both,y explicit
	] coordinates {
		(10, 4)  +- (2.2804, 2.2804) %
		(20,6.25)  +- (3.1125, 3.1125) % 
		(50,24)  +- (7.2388, 7.2388) %
		(100, 52.8)  +- (6.3687, 6.3687) %
	};
	\end{axis}
	\end{tikzpicture}
	\hspace{-1.2em}
	\begin{tikzpicture} % ER-2, Exponental
	\begin{axis}[
	xlabel = {$d$}, xmin = 5, xmax = 105, xtick = {20, 40, 60, 80, 100}, xlabel shift = -4, 
	ymin = 0, ymax = 100, ytick = {20, 40, 60, 80}, yticklabel = \empty, ymajorgrids=true
	]
	\addplot+ [
	mark = o, very thick, 
	error bars/.cd,
	y dir= both,y explicit,
	] coordinates {
		(10, 4.8)  +- (1.833, 1.833) %
		(20, 7.6667)  +- (3.3992, 13.3993) %
		(50, 26.2)  +- (5.2307, 5.2307) %
		(100, 72)  +- (13.1605, 13.1605) %
	};
	
	\addplot+ [
	mark = square, very thick, 
	error bars/.cd,
	y dir= both,y explicit,
	] coordinates {
		(10, 4.2)  +- (2.1354, 2.1354) %
		(20, 7)  +- (1.6733, 1.6733) %
		(50, 25.2)  +- (9.1082, 9.1082) %
		(100, 73.2)  +- (9.8265, 9.8265) %
	};
	\end{axis}
	\end{tikzpicture}
	\hspace{-1.2em}
	\begin{tikzpicture} % ER-2, Gumbel
	\begin{axis}[
	xlabel = {$d$}, xmin = 5, xmax = 105, xtick = {20, 40, 60, 80, 100}, xlabel shift = -4, 
	ymin = 0, ymax = 100, ytick = {20, 40, 60, 80}, yticklabel = \empty, ymajorgrids=true
	]
	\addplot+ [
	mark = o, very thick, 
	error bars/.cd,
	y dir= both,y explicit,
	] coordinates {
		(10, 2)  +- (1.2649, 1.2649) %
		(20, 6.4)  +- (3.3226, 3.3226) %
		(50,15.6)  +- (6.4062, 6.4062) %
		(100, 51.25)  +- (10.2072, 10.2072) %
	};
	
	\addplot+ [
	mark = square, very thick, 
	error bars/.cd,
	y dir= both,y explicit,
	] coordinates {
		(10, 1.75)  +- (1.7854, 1.7854) %
		(20, 5.4)  +- (3.6111, 3.6111) %
		(50, 14.8)  +- (5.2688, 5.2688) %
		(100, 49.6)  +- (7.5525, 7.5525) %
	};
	\end{axis}
	\end{tikzpicture}
	\hspace{-1.2em}
	\begin{tikzpicture} % SF-4, Gaussian
	\begin{axis}[
	xlabel = {$d$}, xmin = 5, xmax = 105, xtick = {20, 40, 60, 80, 100}, xlabel shift = -4, 
	ymin = 0, ymax = 100, ytick = {20, 40, 60, 80}, yticklabel = \empty, ymajorgrids=true
	]
	\addplot+ [
	mark = o, very thick, 
	error bars/.cd,
	y dir= both,y explicit,
	] coordinates {
		(10, 1.8)  +- (0.7483, 0.7483) %
		(20, 8)  +- (0, 0) %
		(50, 26)  +- (9.5917, 9.5917) %
		(100, 58.2)  +- (23.3101, 23.3101) %
	};
		
	\addplot+ [
	mark = square, very thick, 
	error bars/.cd,
	y dir= both,y explicit,
	] coordinates {
		(10, 2)  +- (0.6325, 0.6325) %
		(20, 3.25)  +- (2.0463, 2.0463) %
		(50, 19)  +- (12.0167, 12.0167) %
		(100, 51.8)  +- (14.6888, 14.6888) %
	};
	\end{axis}
	\end{tikzpicture}
	\hspace{-1.2em}
	\begin{tikzpicture} % SF-4, Exponental
	\begin{axis}[
	xlabel = {$d$}, xmin = 5, xmax = 105, xtick = {20, 40, 60, 80, 100}, xlabel shift = -4, 
	ymin = 0, ymax = 100, ytick = {20, 40, 60, 80}, yticklabel = \empty, ymajorgrids=true
	]
	\addplot+ [
	mark = o, very thick, 
	error bars/.cd,
	y dir= both,y explicit,
	] coordinates {
		(10, 5.8)  +- (3.8158, 3.8158) %
		(20, 10.2)  +- (4.7917, 4.7917) %
		(50, 42.75)  +- (13.3112, 13.3112) %
		(100, 51.6)  +- (18.9026, 18.9026) %
	};

	\addplot+ [
	mark = square, very thick, 
	error bars/.cd,
	y dir= both,y explicit,
	] coordinates {
		(10, 5)  +- (1, 1) %
		(20, 4.6)  +- (3.3823, 3.3823) %
		(50, 31.4)  +- (24.6301, 24.6301) %
		(100, 40)  +- (8.8137, 8.8137) %
	};
	\end{axis}
	\end{tikzpicture}
	\hspace{-1.2em}
	\begin{tikzpicture} % SF-4, Gumbel
   \begin{axis}[
   xlabel = {$d$}, xmin = 5, xmax = 105, xtick = {20, 40, 60, 80, 100}, xlabel shift = -4, 
   ymin = 0, ymax = 100, ytick = {20, 40, 60, 80}, yticklabel = \empty, ymajorgrids=true
   ]
   \addplot+ [
   mark = o, very thick, 
   error bars/.cd,
   y dir= both, y explicit,
   ] coordinates {
   	(10, 1)  +- (0, 0) %
   	(20, 6.6)  +- (4.6303, 4.6303) %
   	(50, 48.6)  +- (25.5076, 25.5076) %
   	(100, 45.4)  +- (29.4116, 29.4116) %
   };
   
   \addplot+ [
   mark = square, very thick, 
   error bars/.cd,
   y dir= both,y explicit,
   ] coordinates {
   	(10, 2.6)  +- (0.4899, 0.4899) %
   	(20, 1.2)  +- (0, 0) %
   	(50, 28.8)  +- (1.312, 1.312) %
   	(100, 40)  +- (8.8137, 8.8137) %
   };
   \end{axis}
   \end{tikzpicture}

% Figure for accuracy evaluation results on correlation
	\hspace{-0.3em}
	\begin{tikzpicture} % ER-2, Gaussian
	\begin{axis}[
	xlabel = {$d$}, xmin = 5, xmax = 105, xtick = {20, 40, 60, 80, 100}, xlabel shift = -4, 
	ylabel = {Coefficients}, 	ymin = 0.55, ymax = 1.05, ytick = {0.6, 0.7, 0.8, 0.9, 1.0}, ylabel shift = -4, ymajorgrids=true
	]
	\addplot+ [
	mark = o, very thick, 
	error bars/.cd,
	y dir= both,y explicit,
	] coordinates {
		(10,  0.9910)  +- (0.0064, 0.0064) %
		(20,  0.9227)  +- (0.0472, 0.0472) %
		(50,  0.9019)  +- (0.0562, 0.0562) %
		(100, 0.8960)  +- (0.0487, 0.0487) %
	};
	\end{axis}
	\end{tikzpicture}
	\hspace{-1.2em}
	\begin{tikzpicture} % ER-2, Exponental
	\begin{axis}[
	xlabel = {$d$}, xmin = 5, xmax = 105, xtick = {20, 40, 60, 80, 100}, xlabel shift = -4, 
		ymin = 0.55, ymax = 1.05, ytick = {0.6, 0.7, 0.8, 0.9, 1.0}, yticklabel = \empty, ylabel shift = -4, ymajorgrids=true
	]
	\addplot+ [
	mark = o, very thick,
	error bars/.cd,
	y dir= both,y explicit,
	] coordinates {
		(10, 0.9091)  +- (0.0709, 0.0709) %
		(20, 0.8562)  +- (0.0574, 0.0574) %
		(50, 0.9022)  +- (0.0839, 0.0839) %
		(100, 0.8995)  +- (0.0731, 0.0731) %
	};
	\end{axis}
	\end{tikzpicture}
	\hspace{-1.2em}
	\begin{tikzpicture} % ER-2, Gumbel
	\begin{axis}[
	xlabel = {$d$}, xmin = 5, xmax = 105, xtick = {20, 40, 60, 80, 100}, xlabel shift = -4, 
		ymin = 0.55, ymax = 1.05, ytick = {0.6, 0.7, 0.8, 0.9, 1.0}, yticklabel = \empty, ylabel shift = -4, ymajorgrids=true
	]
	\addplot+ [
	mark = o, very thick, 
	error bars/.cd,
	y dir= both,y explicit,
	] coordinates {
		(10, 0.9248)  +- (0.0505, 0.0505) %
		(20, 0.8870)  +- (0.0966, 0.0966) %
		(50, 0.9259)  +- (0.0407, 0.0407) %
		(100, 0.9124)  +- (0.0752, 0.0752) %
	};
	\end{axis}
	\end{tikzpicture}
	\hspace{-1.2em}
	\begin{tikzpicture} % SF-4, Gaussian
	\begin{axis}[
	xlabel = {$d$}, xmin = 5, xmax = 105, xtick = {20, 40, 60, 80, 100}, xlabel shift = -4, 
	ymin = 0.55, ymax = 1.05, ytick = {0.6, 0.7, 0.8, 0.9, 1.0}, yticklabel = \empty, ylabel shift = -4, ymajorgrids=true
	] 
	\addplot+ [
	mark = o, very thick, 
	error bars/.cd,
	y dir= both,y explicit,
	]  coordinates {
		(10, 0.9843)  +- (0.0108, 0.0108) %
		(20, 0.9475)  +- (0, 0) %
		(50, 0.8201)  +- (0.1768, 0.1768) %
		(100, 0.8056)  +- (0.1888, 0.1888) %
	};	
	\end{axis}
	\end{tikzpicture}
	\hspace{-1.2em}
	\begin{tikzpicture} % SF-4, Exponental
	\begin{axis}[
	xlabel = {$d$}, xmin = 5, xmax = 105, xtick = {20, 40, 60, 80, 100}, xlabel shift = -4, 
		ymin = 0.55, ymax = 1.05, ytick = {0.6, 0.7, 0.8, 0.9, 1.0}, yticklabel = \empty, ylabel shift = -4, ymajorgrids=true
	]
	\addplot+ [
	mark = o, very thick, 
	error bars/.cd,
	y dir= both,y explicit,
	]  coordinates {
		(10, 0.9246)  +- (0.0646, 0.0646) %
		(20, 0.8937)  +- (0, 0) %
		(50, 0.8394)  +- (0.0767, 0.0767) %
		(100, 0.8740)  +- (0. 0847, 0.0847) %
	};	
	\end{axis}
	\end{tikzpicture}
	\hspace{-1.2em}
	\begin{tikzpicture} % SF-4, Gumbel
	\begin{axis}[
	xlabel = {$d$}, xmin = 5, xmax = 105, xtick = {20, 40, 60, 80, 100}, xlabel shift = -4, 
		ymin = 0.55, ymax = 1.05, ytick = {0.6, 0.7, 0.8, 0.9, 1.0}, yticklabel = \empty, ylabel shift = -4, ymajorgrids=true
	]
	\addplot+ [
	mark = o, very thick, 
	error bars/.cd,
	y dir= both,y explicit,
	]  coordinates {
		(10, 0.9351)  +- (0.0592, 0.0592) %
		(20, 0.8857)  +- (0.0709, 0.0709) %
		(50, 0.8166)  +- (0.0878, 0.0878) %
		(100, 0.8685)  +- (0.0301, 0.0301) %
	};	
	\end{axis}	
	\end{tikzpicture}

\hspace{-0.65em}
\begin{tikzpicture} % ER-2, Gaussian
\begin{semilogyaxis}[
xlabel = {$d$}, xmin = 0, xmax = 520, xtick = {100, 200, 500}, xlabel shift = -4, log basis y = {10},
ylabel = {Time (Sec)}, ymin = 7, ymax = 28000, ytick = {10, 100, 1000, 10000}, ylabel shift = -5, ymajorgrids=true
]
\addplot+ [
mark = o, very thick, 
error bars/.cd, 
y dir= both,y explicit,
] coordinates {
	(10,  11.3324)  +- (0, 0) 
	(20, 17.6329)  +- (0, 0) 
	(50, 29.2313)  +- (0, 0)
	(100, 62.9974)  +- (0, 0)
	(200, 127.1861)  +- (0, 0)
	(500, 571.6450)  +- (0, 0)
};
\addplot+ [
mark = square, very thick, 
error bars/.cd,
y dir= both,y explicit
] coordinates {
	(10,  102.2197)  +- (0, 0) 
    (20, 124.4840)  +- (0, 0) 
    (50, 189.0675)  +- (0, 0)
    (100, 373.2211)  +- (0, 0)
    (200, 1213.0571)  +- (0, 0)
    (500, 8275.5216)  +- (0, 0)
};
\end{semilogyaxis}
\end{tikzpicture}
\hspace{-1.2em}
\begin{tikzpicture} % ER-2, Exponental
\begin{semilogyaxis}[
xlabel = {$d$}, xmin = 0, xmax = 520, xtick = { 100, 200, 500}, xlabel shift = -4, log basis y = {10},
ymin = 7, ymax = 28000, ytick = {10, 100, 1000, 10000}, yticklabel = \empty, ymajorgrids=true
]
\addplot+ [
mark = o, very thick, 
error bars/.cd, 
y dir= both,y explicit,
] coordinates {
	(10,  16.1614)  +- (0, 0) 
	(20, 18.7431)  +- (0, 0) 
	(50, 33.8502)  +- (0, 0)
	(100, 64.2954)  +- (0, 0)
	(200, 163.4075)  +- (0, 0)
	(500, 694.3392)  +- (0, 0)
};
\addplot+ [
mark = square, very thick, 
error bars/.cd,
y dir= both,y explicit
] coordinates {
	(10,  106.6996)  +- (0, 0) 
	(20, 137.0397)  +- (0, 0) 
	(50, 208.3243)  +- (0, 0)
	(100, 640.4315)  +- (0, 0)
	(200, 1330.2653)  +- (0, 0)
	(500, 8725.7048)  +- (0, 0)
};
\end{semilogyaxis}
\end{tikzpicture}
\hspace{-1.2em}
\begin{tikzpicture} % ER-2, Gumbel
\begin{semilogyaxis}[
xlabel = {$d$}, xmin = 0, xmax = 520, xtick = {100, 200, 500}, xlabel shift = -4, log basis y = {10},
ymin = 7, ymax = 28000, ytick = {10, 100, 1000, 10000}, yticklabel = \empty, ymajorgrids=true
]
\addplot+ [
mark = o, very thick, 
error bars/.cd, 
y dir= both,y explicit,
] coordinates {
	(10,  20.3721)  +- (0, 0) 
	(20, 21.8578)  +- (0, 0) 
	(50, 34.7230)  +- (0, 0)
	(100, 72.2464)  +- (0, 0)
	(200, 184.3296)  +- (0, 0)
	(500, 592.9941)  +- (0, 0)
};
\addplot+ [
mark = square, very thick, 
error bars/.cd,
y dir= both,y explicit
] coordinates {
	(10,  105.5173)  +- (0, 0) 
	(20, 131.9504)  +- (0, 0) 
	(50, 209.0620)  +- (0, 0)
	(100, 471.1314)  +- (0, 0)
	(200, 1472.1794)  +- (0, 0)
	(500, 8698.1403)  +- (0, 0)
};
\end{semilogyaxis}
\end{tikzpicture}
\hspace{-1.2em}
\begin{tikzpicture} % SF-4, Gaussian
\begin{semilogyaxis}[
xlabel = {$d$}, xmin = 0, xmax = 520, xtick = { 100, 200, 500}, xlabel shift = -4, log basis y = {10},
ymin = 7, ymax = 28000, ytick = {10, 100, 1000, 10000}, yticklabel = \empty, ymajorgrids=true
]
\addplot+ [
mark = o, very thick, 
error bars/.cd, 
y dir= both,y explicit,
] coordinates {
	(10,  16.3669)  +- (0, 0) 
	(20, 17.8526)  +- (0, 0) 
	(50, 35.8852)  +- (0, 0)
	(100, 65.5569)  +- (0, 0)
	(200, 223.8900)  +- (0, 0)
	(500, 1220.5243)  +- (0, 0)
};
\addplot+ [
mark = square, very thick, 
error bars/.cd,
y dir= both,y explicit
] coordinates {
	(10,  105.9523)  +- (0, 0) 
	(20, 125.1070)  +- (0, 0) 
	(50, 216.4841)  +- (0, 0)
	(100, 342.0933)  +- (0, 0)
	(200, 1648.4166)  +- (0, 0)
	(500, 9231.1743)  +- (0, 0)
};
\end{semilogyaxis}
\end{tikzpicture}
\hspace{-1.2em}
\begin{tikzpicture} % SF-4, Exponental
\begin{semilogyaxis}[
xlabel = {$d$}, xmin = 0, xmax = 520, xtick = { 100, 200, 500}, xlabel shift = -4, log basis y = {10},
ymin = 7, ymax = 28000, ytick = {10, 100, 1000, 10000}, yticklabel = \empty, ymajorgrids=true
]
\addplot+ [
mark = o, very thick, 
error bars/.cd, 
y dir= both,y explicit,
] coordinates {
	(10,  19.4405)  +- (0, 0) 
	(20, 26.1188)  +- (0, 0) 
	(50, 35.6885)  +- (0, 0)
	(100, 72.0892)  +- (0, 0)
	(200, 203.0217)  +- (0, 0)
	(500, 1118.8014)  +- (0, 0)
};
\addplot+ [
mark = square, very thick, 
error bars/.cd,
y dir= both,y explicit
] coordinates {
	(10,  115.1225)  +- (0, 0) 
	(20, 143.4021)  +- (0, 0) 
	(50, 259.9929)  +- (0, 0)
	(100, 507.1558)  +- (0, 0)
	(200, 1883.9273)  +- (0, 0)
	(500, 9178.3159)  +- (0, 0)
};
\end{semilogyaxis}
\end{tikzpicture}
\hspace{-1.2em}
\begin{tikzpicture} % SF-4, Gumbel
\begin{semilogyaxis}[
xlabel = {$d$}, xmin = 0, xmax = 520, xtick = {100, 200, 500}, xlabel shift = -4, log basis y = {10},
ymin = 7, ymax = 28000, ytick = {10, 100, 1000, 10000}, yticklabel = \empty, ymajorgrids=true
]
\addplot+ [
mark = o, very thick, 
error bars/.cd, 
y dir= both,y explicit,
] coordinates {
	(10,  17.6187)  +- (0, 0) 
	(20, 22.3945)  +- (0, 0) 
	(50, 39.9701)  +- (0, 0)
	(100, 73.8598)  +- (0, 0)
	(200, 241.2734)  +- (0, 0)
	(500, 1354.8014)  +- (0, 0)
};
\addplot+ [
mark = square, very thick, 
error bars/.cd,
y dir= both,y explicit
] coordinates {
	(10,  117.9441)  +- (0, 0) 
	(20, 143.2404)  +- (0, 0) 
	(50, 250.4158)  +- (0, 0)
	(100, 427.6269)  +- (0, 0)
	(200, 2074.1816)  +- (0, 0)
	(500, 9025.6731)  +- (0, 0)
};
\end{semilogyaxis}
\end{tikzpicture}
\vspace{-1em}
\caption{\normalsize Evaluation results on artificial benchmark data.}
\vspace{-1em}
\label{fig: artiexp-acc}
\end{figure*}

%% file: figs-sca.tex
\pgfplotsset{width = 5cm, height = 4cm}
\pgfplotsset{compat = 1.15}

% Figure for scalability test
\begin{figure*}
\ref{realleg}
\centering

\begin{tikzpicture} % Movielens
\begin{semilogyaxis}[
xlabel = {Execution Time (h)}, xmin = -2, xmax = 95, 
xtick={0, 20, 40, 60, 80},
xlabel shift = -4, log basis y = {10},
ylabel = {Constraint Value}, ymin = 1e-9, ymax = 9e-1, ytick = {1e-8, 1e-6, 1e-4, 1e-2}, ylabel shift = -2, ymajorgrids=true,
title = \textsl{Movielens}, 
	legend style = {draw= none},
	legend columns=-1,
    legend entries={\textsf{$h(\W)$ \quad \quad \quad \quad},\textsf{$\uppdelta(\W)$}},
    legend to name = realleg,
]

\addplot+ [
mark = o
] coordinates {
(0, 9.34296E-07)
(1.616666667, 7.64702E-06)
(3.633333333, 9.65281E-06)
(6, 8.88967E-06)
(7.666666667, 1.27344E-05)
(11.33333333, 1.11351E-05)
(14.08333333, 8.65067E-06)
(16.83333333, 5.71953E-06)
(19.63333333, 5.60955E-06)
(22.53333333, 5.6896E-06)
(25.38333333, 5.64608E-06)
(28.23333333, 5.67976E-06)
(31.1, 5.19829E-06)
(33.96666667, 4.39867E-06)
(36.85, 3.53188E-06)
(39.7, 3.29759E-06)
(42.65, 3.08238E-06)
(45.51666667, 2.26399E-06)
(48.33333333, 1.892E-06)
(51.2, 1.18362E-06)
(54.05, 5.51888E-07)
(56.85, 3.94519E-07)
(59.71666667, 2.9793E-07)
(62.46666667, 1.82841E-07)
(65.26666667, 1.03906E-07)
(68.05, 6.86931E-08)
(70.88333333, 6.16762E-08)
(73.71666667, 5.39088E-08)
(76.73333333, 1.78338E-08)
(79.78333333, 3.78161E-08)
(82.81666667, 2.10086E-08)
(85.66666667, 4.45678E-08)
(88.33333333, 1.40267E-08)
(89.36666667, 8.64264E-09)
};

\addplot+ [
mark=square
] coordinates {
(0, 0.315610585)
(1.616666667, 0.322289077)
(3.633333333, 0.277035502)
(6, 0.255780069)
(7.666666667, 0.242455883)
(11.33333333, 0.253893761)
(14.08333333, 0.172041907)
(16.83333333, 0.300655363)
(19.63333333, 0.164559289)
(22.53333333, 0.119508601)
(25.38333333, 0.118790119)
(28.23333333, 0.09275861)
(31.1, 0.140480622)
(33.96666667, 0.098782896)
(36.85, 0.08414822)
(39.7, 0.064078651)
(42.65, 0.053250394)
(45.51666667, 0.066858686)
(48.33333333, 0.047212079)
(51.2, 0.045934885)
(54.05, 0.046384368)
(56.85, 0.032719098)
(59.71666667, 0.02856815)
(62.46666667, 0.05678017)
(65.26666667, 0.026938502)
(68.05, 0.020147249)
(70.88333333, 0.031369347)
(73.71666667, 0.022398656)
(76.73333333, 0.025993967)
(79.78333333, 0.042645715)
(82.81666667, 0.0186707)
(85.66666667, 0.032625861)
(88.33333333, 0.015927197)
(89.36666667, 0.013537139)
};

\end{semilogyaxis}
\end{tikzpicture}
\hspace{3em}
\begin{tikzpicture} % Appdata-I: http
\begin{semilogyaxis}[
xlabel = {Execution Time (h)}, xmin = -2, xmax = 70, 
xtick={0, 20, 40, 60},
xlabel shift = -4, log basis y = {10}, ylabel = {Constraint Value}, 
ymin = 1e-9, ymax = 9e-1, ytick = {1e-8, 1e-6, 1e-4, 1e-2}, ylabel shift = -5, ymajorgrids=true,
title = \textsl{App-Security}
]

\addplot+ [
mark = o
] coordinates {
	(0, 2.92E-05)
	(4.866666667, 2.92E-07)
	(9.033333333, 1.23E-06)
	(13.1, 1.55E-06)
	(17.16666667, 1.47E-06)
	(21.68333333, 1.07E-06)
	(26.18333333, 1.10E-06)
	(30.71666667, 8.95E-07)
	(35.33333333, 6.85E-07)
	(39.95, 4.05E-07)
	(45.26666667, 2.36E-07)
	(49.58333333, 1.13E-07)
	(54.55, 5.42E-08)
	(59.45, 2.36E-08)
	(64.4, 1.08E-08)
	(67.15, 9.39E-09)
};

\addplot+ [
mark=square
] coordinates {
	(0, 0.22257648)
	(4.866666667, 0.209961532)
	(9.033333333, 0.199583571)
	(13.1, 0.211040756)
	(17.16666667, 0.178235831)
	(21.68333333, 0.146786125)
	(26.18333333, 0.130727494)
	(30.71666667, 0.131002724)
	(35.33333333, 0.114105912)
	(39.95, 0.095508996)
	(45.26666667, 0.077794568)
	(49.58333333, 0.068326261)
	(54.55, 0.05136505)
	(59.45, 0.055351894)
	(64.4, 0.053069682)
	(67.15, 0.050152508)
};

\end{semilogyaxis}
\end{tikzpicture}
\hspace{3em}
\begin{tikzpicture} % Appdata-II: Alimama
\begin{semilogyaxis}[
xlabel = {Execution Time (h)}, xmin = 0, xmax = 7, 
xtick={0, 1.5, 3, 4.5, 6},
xlabel shift = -4, log basis y = {10}, ylabel = {Constraint Value}, 
ymin = 1e-9, ymax = 9e-1, ytick = {1e-8, 1e-6, 1e-4, 1e-2}, ylabel shift = -5, ymajorgrids=true,
title = \textsl{App-Recom}
]

\addplot+ [
mark = o
] coordinates {
	(0, 1.08E-06)
	(0.1, 4.17E-06)
	(0.233333333, 6.41E-06)
	(0.416666667, 5.79E-06)
	(0.616666667, 6.13E-06)
	(0.833333333, 5.72E-06)
	(1.066666667, 3.64E-06)
	(1.316666667, 3.42E-06)
	(1.566666667, 2.46E-06)
	(1.816666667, 1.71E-06)
	(2.1, 1.45E-06)
	(2.416666667, 8.35E-07)
	(2.666666667, 7.32E-07)
	(2.983333333, 6.44E-07)
	(3.266666667, 2.55E-07)
	(3.566666667, 1.85E-07)
	(3.883333333, 1.18E-07)
	(4.216666667, 1.01E-07)
	(4.533333333, 7.67E-08)
	(4.866666667, 6.36E-08)
	(5.2, 5.17E-08)
	(5.333333333, 4.80E-08)
	(5.533333333, 3.27E-08)
	(5.866666667, 2.76E-08)
	(6.183333333, 1.49E-08)
	(6.5, 8.57E-09)
};

\addplot+ [
mark=square
] coordinates {
(0, 0.156247079)
(0.1, 0.1162818)
(0.233333333, 0.110737585)
(0.416666667, 0.098189734)
(0.616666667, 0.07685481)
(0.833333333, 0.073969238)
(1.066666667, 0.074742243)
(1.316666667, 0.066548757)
(1.566666667, 0.056684744)
(1.816666667, 0.041568402)
(2.1, 0.047659811)
(2.416666667, 0.06175046)
(2.666666667, 0.043060482)
(2.983333333, 0.03807541)
(3.266666667, 0.039924886)
(3.566666667, 0.029093713)
(3.883333333, 0.027766079)
(4.216666667, 0.016493801)
(4.533333333, 0.017879732)
(4.866666667, 0.013938545)
(5.2, 0.011187371)
(5.333333333, 0.012462456)
(5.533333333, 0.01098)
(5.866666667, 0.01032)
(6.183333333, 0.00957)
(6.5, 0.00903)
};

\end{semilogyaxis}
\end{tikzpicture}
\vspace{-1em}
\caption{\normalsize Scalability test of \algo algorithm.}
\label{fig: artiexp-sca}
\vspace{-1em}
\end{figure*}
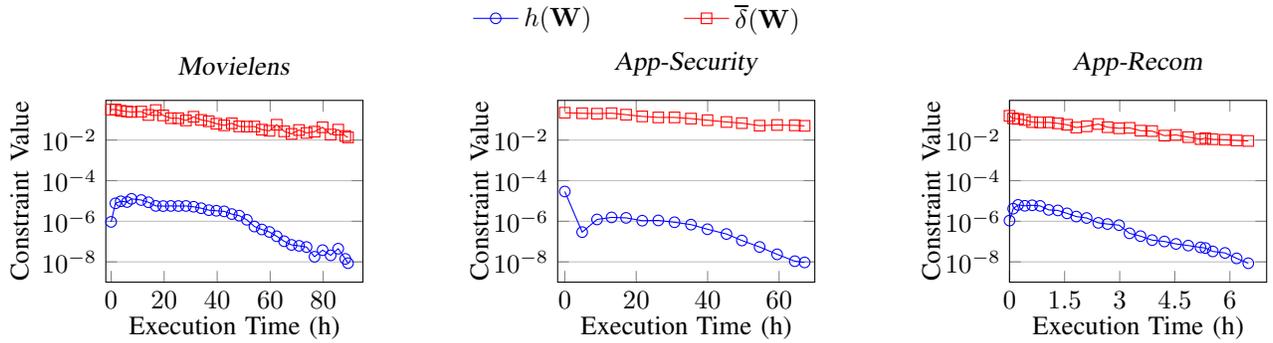

%% file: application.tex
% !TeX spellcheck = en_US

\section{Applications}

In this section, we demonstrate several different types of applications to show the effectiveness of our \textsc{LEAST} method.
Section~VI-A describes how \textsc{LEAST} is used in the production system of ticket booking service.
Section~VI-B describes the usage of \textsc{LEAST} for large-scale gene expression data analysis.
Section~VI-C presents a case study for applying \textsc{LEAST} in the context explainable recommendation system.

\subsection{Monitoring the Flight Ticket Booking Business}

\noindent \underline{\textbf{Scenario Description.}}
Alibaba owns Fliggy (Fei Zhu), a flight ticket booking service which serves several millions of customers every day. It runs in a distributed system of considerable size and complexity, both  growing with each passing day. The system is connected to wide range of interfaces to 
1) a variety of airlines;
2)  large intermediary booking systems such as Amadeus or Travelsky;
3) different smaller travel agents; 4) other internal systems in Alibaba
and etc. We use the term fare source to reflect through which channel a ticket is booked.
The booking process of each flight ticket consists of four essential steps: 1) query and confirm seat availability; 
2) query and confirm price; 3) reserve ticket; 
and 4) payment and final confirmation. 
Each step involved one or several API requests of some interfaces stated above. All this information is logged for monitoring.

Due to many unpredictable reasons, e.g., an update or maintenance of an airline's booking system, 
the booking process may fail at any step. To maintain a low rate of booking errors and ensure customers' experience, an automatic monitoring system has been implemented. Whenever there is a sudden increase in booking errors of each step, log data is inspected to identify possible root causes. Once found, operation staff can then manually intervene.

In this situation, the monitoring system must fulfill two goals:
1) It should detect issues as \emph{fast} as possible in order to reduce the possibility of money loss due to unsuccessful booking.
2) It should clearly find the actual \emph{root} cause of a problem, since a failure originating at a single point may easily influence many other factors in the booking system.

\begin{figure}[t]
	\centering
	\includegraphics[width=0.9\linewidth]{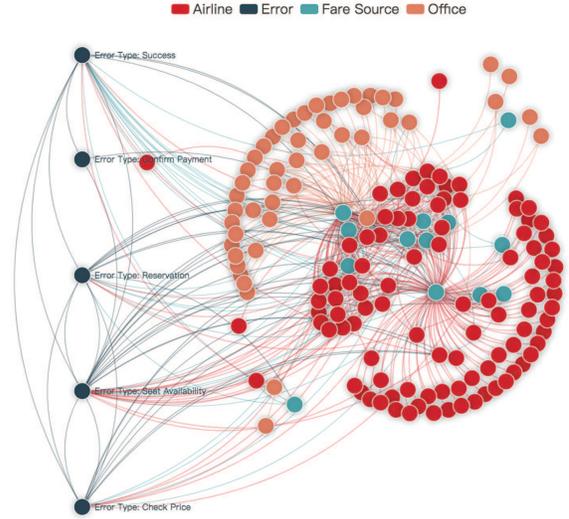}
	\vspace{-0.5em}
	\caption{\normalsize Example of a graph learned from Fliggy booking data with partial nodes.}
	\label{Fig: Feizhu Graph}
	\vspace{-1em}
\end{figure}

\smallskip
\noindent \underline{\textbf{Technical Details.}}
Our \textsc{LEAST} algorithm plays a key role in this monitoring system. It works in the following manner:

1) For every half hour interval, we collect log data $T$ from a window of the latest 24 hours and learn a BN network $G$ using \textsc{LEAST}. The nodes in $G$ include
the four ``error-type nodes" corresponding to the above four booking steps and information on  airlines, fare sources and airports in the ticket booking system.  Fig.~\ref{Fig: Feizhu Graph} illustrates an example of the graph learned by the ticket booking data with partial nodes.

2) For each node $X$ of the four error types, we inspect all  paths $P$ whose destination is $X$. That is, we follow the incoming links of $X$ until we reach a node with no parents. Each $P$ stands for a possible reason causing $X$. To detect whether $P$ is a random coincidence or not, we count the number of occurrences of $P$ in the log data $T$ and $T'$, i.e., the previous window of log data, and perform a statistical test to derive a p-value. Depending on a threshold for the p-value, the entire path $P$ may be reported as an anomaly, with the tail of $P$ likely pinpointing the root cause for the problem.

In general, we have observed that learned BN structure (and weights on edges) remain unchanged across periods, as long as there are no observed anomalies leading to error rate increases. However, in other cases, we have found that relatively large increases in booking error rates are often accompanied by the appearance of new links pointing to the error nodes.

\begin{table*}[t]
	\centering
	\caption{Flight ticket booking example cases identified by \textsc{LEAST}.}
	\vspace{-0.5em}
	\label{tab:feizhu}
	\begin{tabular}{cll}
		\hline
		\textbf{Date} & \textbf{Identified Anomaly Path of Root Cause} & \textbf{Explainable Events} \\ 
		\hline
		2019-11-19& Error in Step 3 $\leftarrow$  Fare sources 3,9,16 $\leftarrow$ Airline AC & \makecell[l]{Air Canada booking system unscheduled maintenance, which \\ creates problems for a variety of different fare sources} \\  \hline
		2019-12-05& Error in Step 3 $\leftarrow$  Airline SL $\leftarrow$ Agent Office BKKOK275Q & Inaccurate data for Airline SL from Amadeus\\ \hline
		2019-12-09& \makecell[l]{Error in Step 3 $\leftarrow$ Error code P-1005-16-001:-41203 \\ Error in Step 3 $\leftarrow$ Airline MU $\leftarrow$ Fare source 5} & Problem caused by internal system deployment \\ \hline
		2020-01-23& Error in Step 1 $\leftarrow$ Arrival city WUH & Lock-down of Wuhan City and many flights are cancelled \\ \hline
		2020-02-15/20/28 & Error in Step 1 $\leftarrow$ Arrival city BKK &
		\makecell[l]{Australia extended travel ban from China and \\ passengers sought Bangkok as a transfer point} \\ \hline
		2020-02-24 & \makecell[l]{Error in Step 1 $\leftarrow$ Departure city SEL \\ Error in Step 1 $\leftarrow$ Airline MU}  & COVID-19 broke out in South Korea\\
		\hline
	\end{tabular}
	\vspace{-2em}
\end{table*}

\smallskip
\noindent \underline{\textbf{Effects and Results.}}
Owing to the high efficiency of our \textsc{LEAST} method, the analysis task could be finished in $2$--$3$ minutes of each run.
Therefore, the operation staff can take immediate actions to resolve the problems, rather than waiting for several hours or days for domain expert analysis and feedback as before. After deploying our  \textsc{LEAST}-based monitoring system, the \emph{first-attempt success booking rate}, a metric evaluating the percentage of all tickets having been successfully booked upon the first try, has been improved.

\begin{figure}[t]
\centering
\begin{tikzpicture}
 \pie[radius=2]{42/\small{external systems}, 3/\small{airline}, 10/\small{travel agent},  3/\small{intermediary interfaces}, 
 	39/\small{unpredictable events},  3/\small{false alarms}}
\end{tikzpicture}
\vspace{-2em}
\caption{\normalsize Real-world evaluation results.}
\label{fig:feizhu}
\vspace{-2em}
\end{figure}
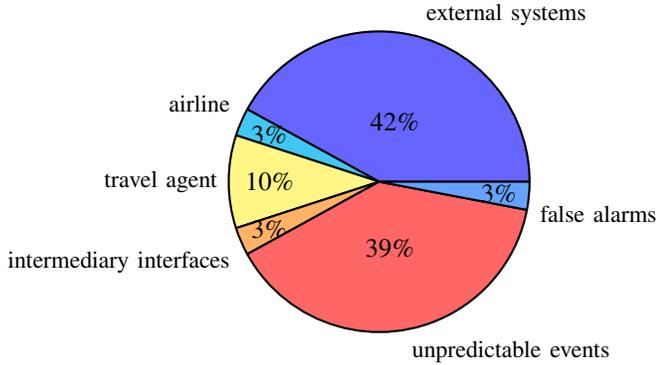

To evaluate our \textsc{LEAST}-based monitoring system, we have compared the output of \textsc{LEAST} for a period of several weeks with results obtained manually from domain experts. Fig.~\ref{fig:feizhu} reports the results of this evaluation. Clearly, only $3\%$ of the reported cases are found to be false positives but $97\%$ of them are true positives. Among them, $42\%$ of the reported problems occur relatively frequently due to problems with some external systems. $3\%$, $3\%$ and $10\%$ of the problems are correctly associated to issues caused by airlines, intermediary interfaces and travel agents, respectively, which has been 
verified manually. The remaining $39\%$ were related to actual problems, the root causes of which remained unknown even after manual inspection and analysis, which can be due to a variety of highly unpredictable events (such as cabin or route adjustments from airlines and sudden weather changes).

Table~\ref{tab:feizhu} reports some examples of observed anomalies with explainable events. Some of these problems were ultimately caused by technical issues with an airline or intermediary interface. Other issues listed in Table~\ref{tab:feizhu} were caused by effects of the COVID-19 pandemic, such as sudden travel restrictions. While these events are not directly observable from the log data, our method was able to successfully detect their effects in our business.

\smallskip
\noindent \underline{\textbf{Summary.}}
Overall, our LEAST-based anomaly detection and root cause analysis system has greatly reduced the amount of manual labor for operation staff and improves the user experience of its service. We also plan to deploy \textsc{LEAST} into more similar applications.

\subsection{Gene Expression Data Analysis}

We further apply our \textsc{LEAST} for gene expression data analysis, a traditional application of structure learning for BN~\cite{lee2019scaling, scanagatta2019survey}. We use the three well-known datasets \textsl{Sachs}~\cite{bnrepo}, \textsl{E. Coli}~\cite{schaffter2011genenetweaver} and \textsl{Yeast}~\cite{schaffter2011genenetweaver}.
In each dataset, each node is a gene and each sample represents the normalized expression level values of some genes in a biological bench experiment. The underlying networks are commonly referred to as gene regulatory networks. We have the ground truth on the BN structure of them.

We run our \textsc{LEAST} method and the state-of-the-art \textsc{NOTEARS} method on a GPU. The evaluation results are reported in Table~\ref{Tab: GeneRes}. Regarding accuracy, we inspect the false direction rate (FDR), true positive rate (TPR), the false positive rate (FPR) and ROC-AUC metrics. 
We observe that the results of \algo appear to be slightly better compared to \textsc{NOTEARS} on all gene datasets. \algo detects $45$ and $104$ more true positive edges than \textsc{NOTEARS} on \textsl{E. Coli} and \textsl{Yeast}, respectively.
Relatively, on \textsl{E. Coli}, \algo improves the $F_1$-score and AUC-ROC by $3.5\%$ and $5.2\%$, respectively. 

We conjecture that this is due to slight numerical instabilities due to the iterative nature of the process for computing $\uppdelta(\W)$, which may eventually cause a higher variance in the final results. However, it appears that this additional source of noise acts as a form of implicit regularization, causing \algo to exhibt higher accuracy in the gene dataset experiments.

On a GPU, in terms of the run time, the speed up ratio appears to be as not as significant as in the CPU case. This is could be due to the high level of parallelism possible in modern GPUs, concealing the speed up effects introduced by the lower complexity matrix operations in \textsc{LEAST}.
However, note that, typical sizes of modern GPU memory does not allow for processing graphs larger than e.g. \textsl{Yeast}.

\begin{table*}[t]
	\centering
	\caption{Properties of real-world datasets in experiments.}
	\vspace{-0.5em}
	\begin{tabular}{ccccccc}
		\hline
		\multirow{2}{*}{\textbf{ Metric}} & \multicolumn{2}{c}{\textbf{\textsl{ Sachs}}} & \multicolumn{2}{c}{\textbf{\textsl{ E. Coli}}} & \multicolumn{2}{c}{\textbf{\textsl{ Yeast}}} \\ \cline{2-7} 
		\specialrule{0em}{2pt}{2pt} 
		& \textsc{NOTEARS} & \textsc{LEAST} & \textsc{NOTEARS} & \textsc{LEAST} & \textsc{NOTEARS} & \textsc{LEAST} \\ \hline
		\specialrule{0em}{2pt}{2pt} 
		\# of Nodes & \multicolumn{2}{c}{  $11$} &  \multicolumn{2}{c}{  $1,565$} & \multicolumn{2}{c}{  $4,441$}  \\  
		\# of Samples & \multicolumn{2}{c}{  $1,000$} &  \multicolumn{2}{c}{  $1,565$} & \multicolumn{2}{c}{  $4,441$}  \\ 
		\# of Exact Edges & \multicolumn{2}{c}{  $17$} &  \multicolumn{2}{c}{  $3,648$} & \multicolumn{2}{c}{  $12,873$}  \\  \hline
		\specialrule{0em}{3pt}{2pt} 
		\# of Predicted Edges &  {$17$} &  {$15$} &  {$164$} &
		{$234$} &  {$560$} &  {$794$}
		\\
		\# of True Positive Edges &  {\textbf{7}} &  {\textbf{7}} &  {$54$} &  {\textbf{99}} & {$307$}  &  {\textbf{411}} \\ \hline
		\specialrule{0em}{2pt}{2pt} 
		FDR &  \textbf{0.353} &   \textbf{0.353} &  $0.146$ &  \textbf{0.103} &  $0.013$ &  \textbf{0.011} \\ 
		TPR &  \textbf{0.412} &  \textbf{0.412} &  $0.047$  &  \textbf{0.066} &  0.044 &  \textbf{0.062}\\ 
		FPR &  $0.263$ &  \textbf{0.211} &  $\mathbf{1.97 \times 10^{-5}}$ &  $\mathbf{1.97 \times 10^{-5}}$ &  $\mathbf{7.11 \times 10^{-7}}$ &  $9.14 \times 10^{-7}$\\ 
		SHD &  $14$ &  \textbf{12} &  $3, 492$ &  \textbf{3,422} &  $12, 311$ &  \textbf{12, 079}\\ 
		$F_1$-score &  $0.412$ &  $\mathbf{0.437}$ &  $0.073$ &  \textbf{0.108} &  0.082 &  \textbf{0.119} \\ 
		AUC-ROC &  $0.925$ &  \textbf{0.947} &  $0.831$ &  \textbf{0.883} &  0.891 &  \textbf{0.919} \\ \hline
		\specialrule{0em}{3pt}{2pt}
		GPU Time (Sec) &  $134$ &  \textbf{53} &  $811$ &  \textbf{415} &  \textbf{6,610} &  $6,930$ \\ \hline
	\end{tabular}
	\label{Tab: Dataset}
	\vspace{-1.5em}
\end{table*}

\subsection{Recommendation Systems: \textsl{Movielens} Case Study}
We provide here a case study using the \textsl{Movielens} dataset to show how structure learning might find an application in the context of recommendation systems. The purpose is to exemplify how structure learning, once it is possible to apply it on large-scale datasets, might play a role in understanding user behavior and building explainable recommendation systems.
In particular, we find that the resulting DAG for this case study resembles item-to-item graphs typically encountered in neighborhood based recommendation systems~\cite{defazio2012graphical}. Yet, since our method is agnostic to the real-world meaning of the random variables $X_i$, using structure learning one could also learn graphs which combine relations between items, users, item features and user features in a straight forward manner.

\begin{table*}[t]
	\centering
	\caption{Examples for top-10 learned edges for \textsl{Movielens}}
	\vspace{-0.5em}
	\begin{tabular}{ccccc}
		\hline
		\textbf{ Link From} & & \textbf{ Link To} & \textbf{ Weight} & \textbf{Remarks} \\ 
		\hline Shrek 2 (2004) & $\rightarrow$ & Shrek (2001) & 0.220 & same series\\
		Raiders of the Lost Ark (1981) & $\rightarrow$ & Star Wars: Episode IV (1977) &  0.178 &  same main actor\\
%The Usual Suspects (1995) & $\longrightarrow$ & The Shawshank Redemption (1994) & 0.177 & \\
		Raiders of the Lost Ark (1981) & $\rightarrow$ & Indiana Jones and the Last Crusade (1989) & 0.159 & same series\\
Harry Potter and the Chamber of Secrets (2002) & 
$\rightarrow$ & Harry Potter and the Sorcerer's Stone (2001) & 0.159 & same series\\
The Maltese Falcon (1941) & $\rightarrow$ & Casablanca (1942) & 0.159 & same period\\
%\tabincell{c}{A Grand Day Out with \\ Wallace and Gromit(1989)} &
%$\longrightarrow$ & \tabincell{c}{Wallace and Gromit: \\ The Wrong Trousers (1993)} & 0.152 & same series\\
Reservoir Dogs (1992) & $\rightarrow$ & Pulp Fiction (1994) & 0.146 & same director\\
North by Northwest (1959) & $\rightarrow$ & Rear Window (1954) & 0.144 & same director\\
Toy Story 2 (1999) & $\rightarrow$ & Toy Story (1995) & 0.144 & same series\\
Spider-Man (2002) & $\rightarrow$ & Spider-Man 2 (2004) & 0.126 & same series\\
Seven (1995) & $\rightarrow$ & The Silence of the Lambs (1991) & 0.126 & same genre\\
%Raiders of the Lost Ark (1981) & $\longrightarrow$ & Back to the Future (1985) & 0.125 & \\
\hline
	\end{tabular}
	\label{Tab: MovielensRes1}
	\vspace{-2em}
\end{table*}
\begin{figure}[t]
	\centering
	\includegraphics[width=0.9\linewidth]{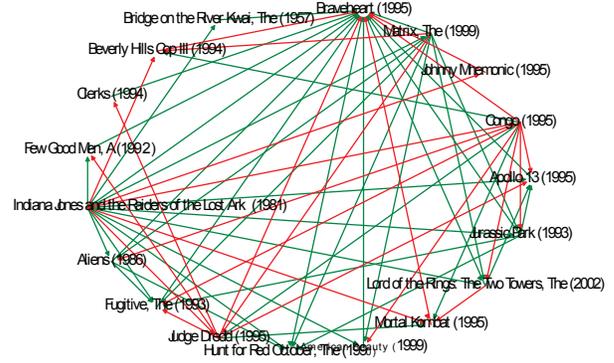}
	\caption{\normalsize Example extracted subgraph: Green and red edges indicate positive and negative learned weights, respectively.}
	\label{Fig: MovielensRes1}
	\vspace{-1.5em}
\end{figure}

A qualitative inspection of the resulting DAG yields interesting insights into the data. We present some of the learned edges with the strongest weights in Table~\ref{Tab: MovielensRes1}. We often find, that links with strong positive weights indicate very similar movies, i.e. movies from the same series, period or director.  

Furthermore, a representative subgraph extracted around the Movie \textit{Braveheart} is presented in Figure \ref{Fig: MovielensRes1}. One may directly interpret the learned structure as follows: given a user's rating for a specific movie, we start at the node corresponding to the movie $i$ and follow outgoing edges to movies $j$, while multiplying the rating with weights for edges $i \rightarrow j$. If resulting values are positive one could predict whether based on the original rating for movie $i$, a user might like (positive value) or dislike the movie $j$. This indicates the possibility of building a simple, yet explainable recommendation system based on the learned graph structure.

We also observe an interesting phenomenon related to the acyclicity constraint: So called ``blockbuster'' or very popular movies, which can be assumed to have been watched by the majority of users, such as e.g. \textit{Star Wars: Episode V} (no outgoing, but 68 incoming links) or \textit{Casablanca} (no outgoing, but 48 incoming links) tend to have a larger number of incoming links than outgoing links. On the other hand, less well known movies or movies indicative of a more specialized taste, such as \textit{The New Land} (with no incoming, but 221 outgoing links), tend to have more outgoing links. 

One possible explanation for this could be given along the thought: Since the majority of users enjoy e.g. \textit{Star Wars}, we do not obtain any interesting insight into a user's taste if we only knew that he or she would like e.g. \textit{Star Wars: Episode V} (no outgoing, but 68 incoming links) or \textit{Casablanca} (no outgoing, but 48 incoming links). On the other hand, enjoying a movie such as the 1972 Swedish movie \textit{The New Land} (with no incoming, but 221 outgoing links) seems to be much more indicative of a user's personal tastes in movies.